\documentclass{article}

\PassOptionsToPackage{numbers, compress}{natbib}

\usepackage[preprint]{wenxiaobai}

\usepackage[utf8]{inputenc} 
\usepackage[T1]{fontenc}    
\usepackage{url}            
\usepackage{booktabs}       
\usepackage{amsfonts}       
\usepackage{nicefrac}       
\usepackage{microtype}      
\usepackage{atbegshi}

\usepackage{adjustbox}
\usepackage{multirow}
\usepackage[table]{xcolor}
\usepackage{caption}
\usepackage{subcaption}
\usepackage{amsmath}
\usepackage{amssymb}
\usepackage{geometry}
\usepackage{wrapfig}
\usepackage{tcolorbox}  
\tcbuselibrary{breakable, skins}
\usepackage{enumitem}
\usepackage{fvextra}
\usepackage{hyperref}       
\usepackage{pifont}

\DefineVerbatimEnvironment{Verbatim}{Verbatim}{breaklines=true, breakanywhere=true}

\hypersetup{
    colorlinks=true,  
    citecolor=black,
    urlcolor=black,   
    linkcolor=black
}

\title{MCP-AgentBench: Evaluating Real-World Language Agent Performance with MCP-Mediated Tools}

\author{
{\bf Zikang Guo\textsuperscript{\rm 1}\thanks{Work done during the internship at Metastone.}\, , Benfeng Xu\textsuperscript{\rm 1,2}, Chiwei Zhu\textsuperscript{\rm 1}, Wentao Hong\textsuperscript{\rm 2}, Xiaorui Wang\textsuperscript{\rm 2}, Zhendong Mao\textsuperscript{\rm 1}\thanks{Corresponding author: Zhendong Mao.}} \\
\textsuperscript{1}University of Science and Technology of China\\
\textsuperscript{2}MetastoneTechnology, Beijing, China \\ 
\texttt{\{gzk170401, benfeng\}@mail.ustc.edu.cn}
}

\begin{document}

\maketitle

\begin{abstract}
The Model Context Protocol (MCP) is rapidly emerging as a pivotal open standard, designed to enhance agent-tool integration and interoperability, and is positioned to unlock a new era of powerful, interconnected, and genuinely utilitarian agentic AI. However, despite MCP's growing adoption, existing benchmarks often fail to capture real-world agent performance within this new paradigm, leading to a distorted perception of their true operational value and an inability to reliably differentiate proficiencies. To bridge this critical evaluation gap, we introduce MCP-AgentBench—a comprehensive benchmark specifically engineered to rigorously assess language agent capabilities in MCP-mediated tool interactions. Core contributions of MCP-AgentBench include: the establishment of a robust MCP testbed comprising 33 operational servers with 188 distinct tools; the development of a benchmark featuring 600 systematically designed queries distributed across 6 distinct categories of varying interaction complexity; and the introduction of MCP-Eval, a novel outcome-oriented evaluation methodology prioritizing real-world task success. Through extensive empirical evaluation of leading language agents, we provide foundational insights. MCP-AgentBench aims to equip the research community with a standardized and reliable framework to build, validate, and advance agents capable of fully leveraging MCP's transformative benefits, thereby accelerating progress toward truly capable and interoperable AI systems.
\end{abstract}

\section{Introduction}

\begin{wrapfigure}{r}{0.45\textwidth}
  \centering
  \includegraphics[width=0.95\linewidth]{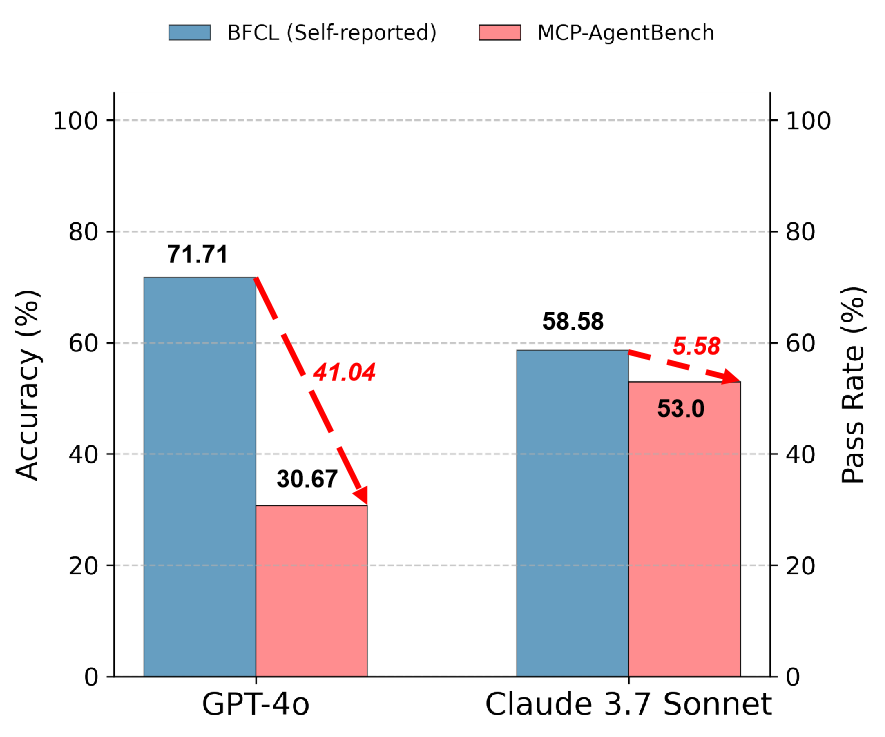}
  \captionof{figure}{BFCL \cite{berkeley-function-calling-leaderboard} vs. MCP-AgentBench}
  \label{fig:bfcl_mcpagentbench_wrap} 
  \vspace{-15pt} 
\end{wrapfigure}

Language agents, leveraging Large Language Models (LLMs) for reasoning and interaction \cite{sumers2023cognitive, wang2024survey, guo2024large, su2024language}, are rapidly emerging as a transformative force in AI.
Their ability to autonomously operate in digital environments, including navigating the web \cite{zhou2023webarena, pan2024webcanvas, xue2025illusion}, controlling applications \cite{xie2024osworld, niu2024screenagent, Claude-computeruse}, and interfacing with software tools \cite{yang2024swe, liu2024large}, signals significant advancements.
However, a primary hurdle to widespread agent utility is achieving effective and scalable interaction with the external world. This involves not only agent-level mastery of complex tool-use workflows \cite{patil2024gorilla, qin2023toolllm, yang2024swe} (from intent understanding to multi-step orchestration) but also resolving the ecosystem-level "M$\times$N integration problem", where M agents must interface with N disparate tools, creating a complex and fragmented landscape that hinders interoperability for cohesive, large-scale agentic systems.

To address these obstacles, the Model Context Protocol (MCP) \cite{mcp} has emerged as a pivotal open standard for AI-tool interactions, providing a universal communication layer and standardized interface to simplify the integration landscape. Crucially, MCP distinguishes itself from traditional function calling \cite{qin2023toolllm, yao2024tau} approaches in several key aspects. Firstly, while function calling typically treats tools as discrete, callable units, MCP is designed to facilitate interaction with a server as a more holistic entity, potentially managing context or state across multiple operations based on the protocol's specifications. Secondly, MCP standardizes the feedback mechanism from the environment, ensuring agents receive richer, more actionable, and consistent responses that go beyond simple function return values, which is vital for robust autonomous operation. Thirdly, unlike paradigms requiring developers to meticulously pre-define tool schemas for the agent, MCP's standardized nature is envisioned to support more dynamic tool discovery and interaction, with servers exposing their capabilities in a universally understandable format, thus reducing bespoke integration efforts and enhancing agent adaptability. By fostering such an open, interoperable, and scalable environment, MCP lays the foundation for a new generation of autonomous and deeply integrated AI experiences, positioning it as a fundamental enabler for powerful, interconnected, and utilitarian agentic AI \cite{hou2025model}.

However, MCP's emergence necessitates an evolution in agent evaluation. Existing function calling benchmarks \cite{berkeley-function-calling-leaderboard, qin2023toolllm, yao2024tau, zhong2025complexfuncbench}, lacking consideration for such standardized interaction protocols, are ill-suited for MCP-mediated activities. Consequently, their failure to accurately measure effectiveness or reliably differentiate capabilities creates a critical evaluation gap, hindering both the development of MCP-based systems and systematic progress in leveraging MCP's transformative benefits. 
For instance, scores from benchmarks like the BFCL \cite{berkeley-function-calling-leaderboard}, which typically evaluate a narrower scope of tool interactions (Figure \ref{fig:bfcl_mcpagentbench_wrap}), can misrepresent practical efficacy.
While real-world user experiences (e.g., on OpenRouter \cite{openrouter}) suggest models such as Claude 3.7 Sonnet \cite{Claude37} can outperform GPT-4o \cite{gpt4o} in complex tool-assisted tasks, their BFCL rankings may significantly diverge. Such discrepancies underscore the limitations of current benchmarks in gauging true agent effectiveness, particularly for standardized, protocol-driven interactions as promoted by MCP, thereby risking misinterpretations of model capabilities.

To bridge this critical evaluation gap, we introduce MCP-AgentBench, a comprehensive benchmark engineered to assess language agent proficiency in MCP-mediated tool use. 
Our benchmark relies on a meticulously developed MCP server testbed, a unified platform with diverse, deployed MCP-compliant servers, providing a realistic and standardized evaluation environment.
This testbed comprises 33 MCP servers offering 188 tools, selected based on stringent criteria of executability, statelessness (to ensure reproducibility in initial evaluations), and text-based interaction.

MCP-AgentBench features systematically categorized queries spanning a spectrum of interaction complexities, from single-server operations to multi-server sequential workflows requiring sophisticated planning and information synthesis. Its data construction follows a rigorous three-stage process: (1) establishing the MCP server testbed; (2) generating diverse, realistic queries via a structured categorization framework that systematically varies interaction complexity (e.g., number of servers, inter-server dependencies, reasoning depth); and (3) creating reference answers using an LLM-assisted, human-verified annotation process.

To evaluate agent performance objectively, we developed MCP-Eval, an LLM-as-a-judge \cite{zheng2023judging} methodology tailored for MCP-AgentBench. 
This approach prioritizes real-world effectiveness by assessing tangible task success rather than rigid adherence to specific execution paths, recognizing that multiple valid solution trajectories often exist for complex problems.

In summary, our main contributions are:

\begin{enumerate}
\item The establishment of a comprehensive MCP server testbed, integrating diverse MCP-compliant servers under a unified system to provide a standardized environment for MCP research and agent development.
\item MCP-AgentBench, a comprehensive benchmark built on this testbed, designed to rigorously evaluate language agent proficiency within the MCP paradigm. It comprises 600 queries across 6 distinct categories, featuring diverse, realistic tasks that probe complex agent-tool interaction patterns, coupled with MCP-Eval, our LLM-as-a-judge methodology that assesses tangible task success.
\item Foundational empirical insights from evaluating leading language agents with MCP-AgentBench, highlighting current capabilities, limitations, and challenges in mastering the MCP framework, thereby offering directions for future advancements within this paradigm.
\end{enumerate}

\begin{figure*}[t]
\centering
\includegraphics[width=0.99\linewidth]{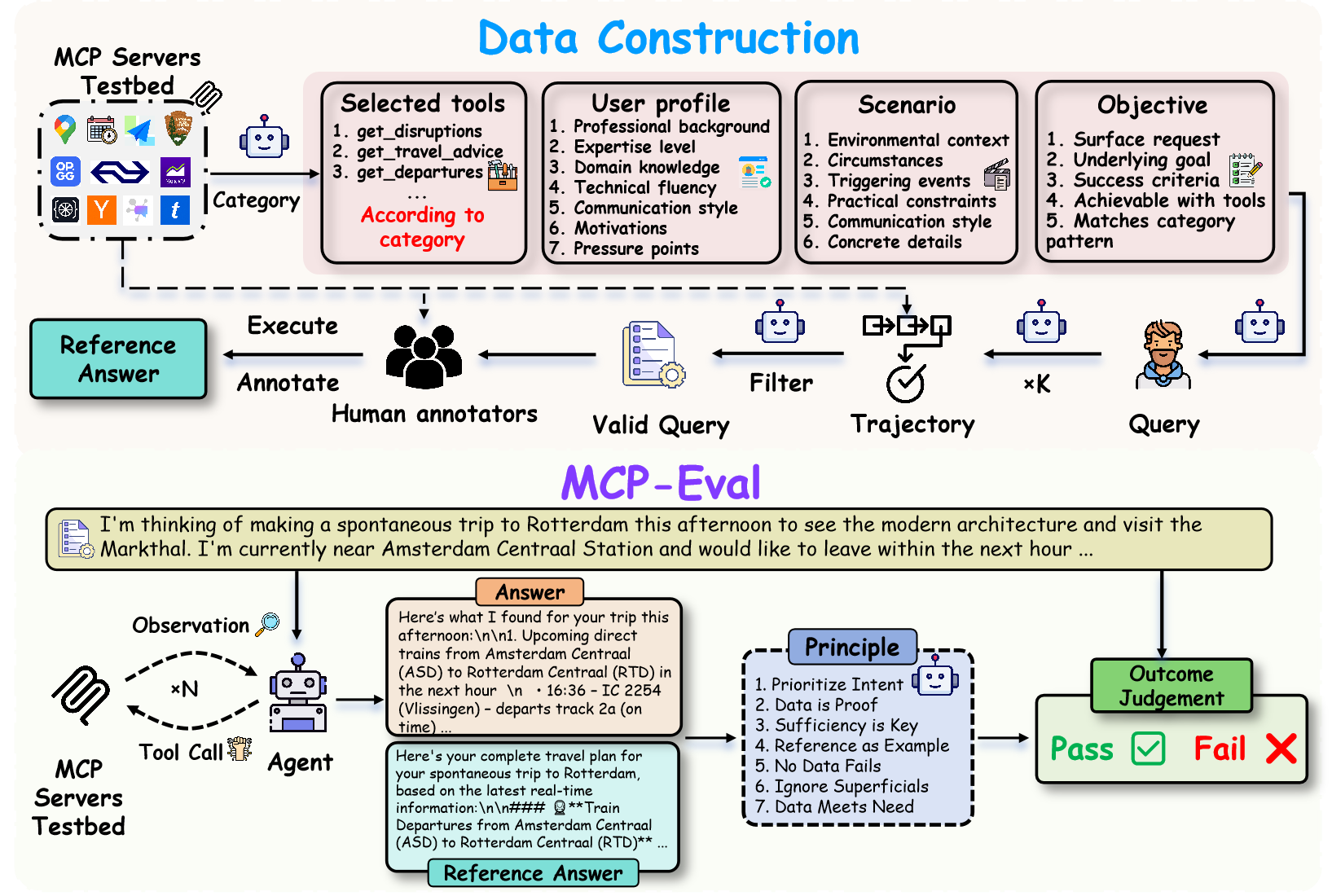}
\caption{The construction and evaluation workflow of MCP-AgentBench. Top (Data Construction): (1) A unified MCP server testbed is established. (2) Diverse, categorized queries are generated with LLM assistance and human verification. (3) Reference answers are produced via an expert-in-the-loop framework. Bottom (MCP-Eval): An LLM-as-a-judge performs an outcome-oriented evaluation of the agent's final answer, yielding a pass/fail score based on task completion.}
\label{fig:data_construction_workflow}
\end{figure*}

\section{MCP-AgentBench}
\label{sec:MCP-AgentBench} 

This section details the development of MCP-AgentBench, our novel benchmark for evaluating agents in MCP environments. We first describe the comprehensive data construction pipeline, which encompasses the establishment of the MCP server testbed, the methodology for query generation, and the process for reference answer annotation. Subsequently, we present key statistics of the resulting dataset and conclude by introducing MCP-Eval, our automated framework for evaluating agent performance on this benchmark.

\subsection{Data Construction Methodology} 
\label{sec:data_construction}

The construction of MCP-AgentBench followed a rigorous, multi-stage methodology, illustrated in Figure~\ref{fig:data_construction_workflow}. This process involved three primary stages: (1) establishing the MCP server testbed, (2) generating diverse user queries, and (3) annotating corresponding reference answers. An LLM (specifically, the Claude 3.7 Sonnet model \cite{Claude37}) provided automated assistance throughout these procedures. To ensure high data quality, executability, and overall reliability, each stage of this process—from server curation and query generation to answer annotation—incorporated a meticulous human-in-the-loop verification procedure. Human annotators validated key aspects and outputs, thereby establishing a robust foundation for the benchmark and subsequent evaluations.

\subsubsection{Establishing the MCP Server Testbed: Rigorous Curation, Deployment, and Integration}
\label{sec:mcp_servers_collection}

The foundational phase in developing MCP-AgentBench was the meticulous construction of a robust and diverse MCP server testbed. This process began with a rigorous curation procedure, involving an extensive survey that initially identified 369 candidate MCP servers. From this large pool, a stringent manual filtering process was applied to select the final set for inclusion. This selective approach was necessitated by the significant challenges inherent in testing and benchmarking a large, heterogeneous server population. Our selection criteria prioritized servers that were: (1) readily executable and stable, (2) stateless in their operational paradigm, and (3) primarily reliant on text-based input and output as defined by the MCP. The focus on text-based interactions stemmed from the complexities associated with the quantitative evaluation of non-textual modalities. Furthermore, stateful servers, which require persistent state across independent benchmark queries, were deemed unsuitable for our framework due to the intricate requirements for managing initial states, tracking temporal changes, and handling state resets, all of which impede reproducible assessment. This meticulous screening, accomplished by a team of three people over seven days, yielded 33 servers that met our stringent criteria.

\begin{figure*}[ht]
\centering
\includegraphics[width=0.86\linewidth]{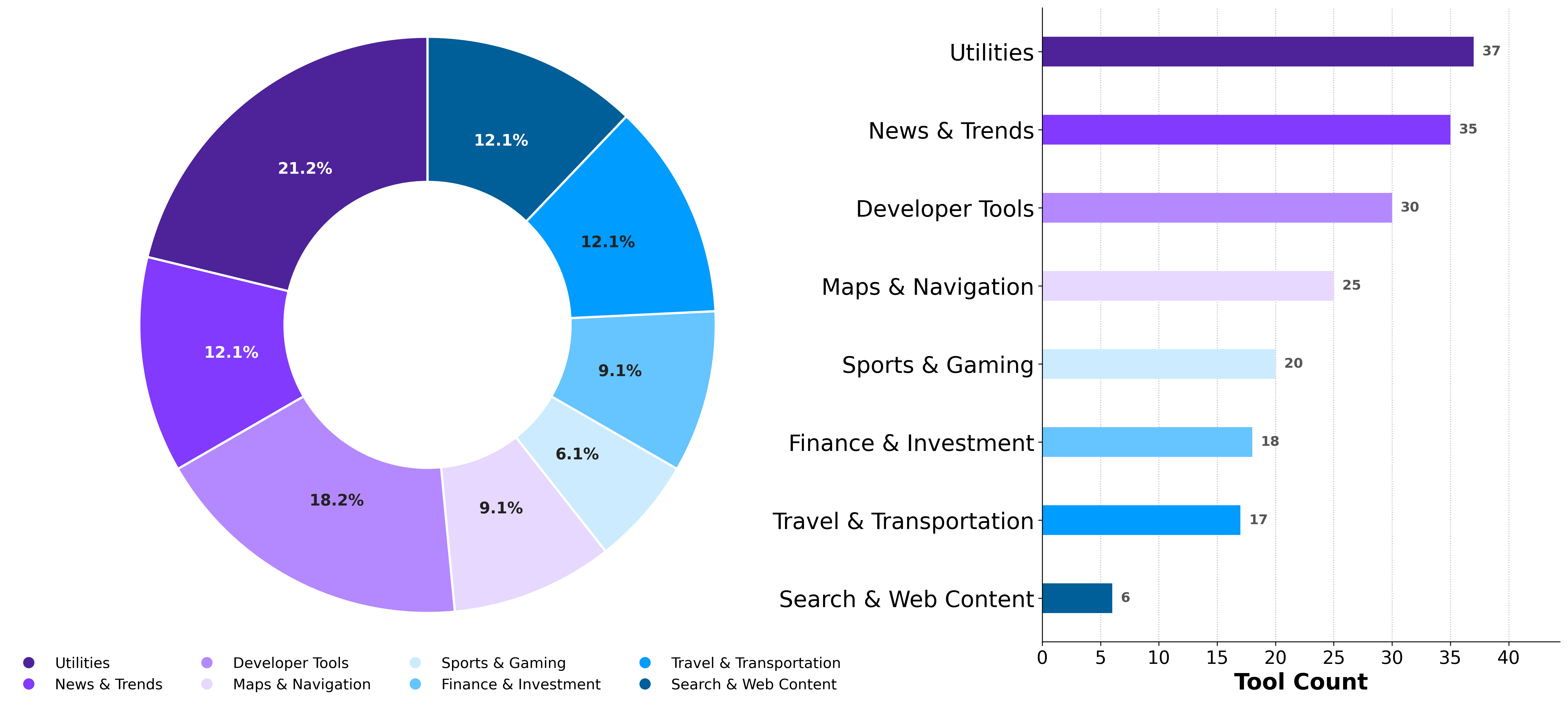}
\caption{Characterization of the MCP server distribution in our testbed. Left: topic-based distribution of the MCP servers; Right: number of tools available per topic across these servers.}
\label{fig:mcp_server_distribution}
\end{figure*}

Following curation, the 33 selected servers underwent careful deployment and configuration in strict accordance with their official operational guidelines. Subsequently, for seamless and standardized agent interaction, all deployed servers were consolidated under a unified invocation interface. This was achieved by leveraging mcprouter \cite{chatmcpmcprouter}, a tool that abstracts server-specific operational idiosyncrasies and facilitates uniform communication protocols. This comprehensive process culminated in an operational testbed comprising 33 distinct MCP servers, which collectively provide access to 188 unique tools, as detailed in Figure~\ref{fig:mcp_server_distribution}.

\subsubsection{Query Generation}
\label{sec:query_generation}

To address the limitations of existing benchmarks in controlling complexity and ensuring realism, we developed an advanced query generation methodology. This methodology employs an LLM to construct test queries that are valid, verifiable, realistic, and diverse. The process unfolds in three core phases. \textbf{First}, we establish a structured categorization framework to systematically vary interaction complexity. \textbf{Second}, we leverage the LLM, guided by a detailed set of instructions, to generate a coherent set of contextual components, including necessary tools, user profiles, background scenarios, and explicit objectives. \textbf{Finally}, these components are synthesized into naturalistic user queries by the LLM, followed by a rigorous human verification process to produce the final, validated queries for our dataset.

\paragraph{Category Definition}
The objective of the first phase is to systematically control and vary the complexity of interaction patterns within our test queries. To this end, we established a categorization framework defined by two primary dimensions: \textbf{server scope} and \textbf{call dependency}. Server scope distinguishes whether task resolution is confined to a single server or requires coordination across multiple servers. Call dependency differentiates tasks based on whether they require a single tool invocation, concurrent invocations of multiple tools (parallel calls), or a sequence of dependent invocations (sequential calls). The combination of these dimensions yields six distinct categories (e.g., \texttt{single-server-single-call} or \texttt{multi-server-sequential-call}), which span a spectrum from simple requests to complex, multi-step workflows.

\paragraph{Contextual Component Generation (\textbf{Phase 2})} 
In the second phase, the LLM generates a set of contextual components tailored to a specific category from our framework. Guided by a comprehensive prompt, the model produces a single, structured output containing four internally consistent elements:

\subparagraph{Tool Selection} The model identifies the specific set of tools from the testbed required to fulfill the user's objective, ensuring the chosen tools are sufficient and produce deterministic, verifiable outputs aligned with the category's complexity.

\subparagraph{User Profile} The model generates a detailed user persona, including professional background and communication style, to ground the query in a realistic context.

\subparagraph{Scenario Description} The model establishes a narrative context that motivates the user's need. Scenarios are calibrated for complexity and include concrete details that naturally yield the necessary parameters for the selected tools.

\subparagraph{User Objective} The model explicitly defines the user's intended goal and verifiable success criteria. A critical constraint is that each objective must be fully achievable using only the tools selected in the same generation step.
This unified generation process ensures that all contextual components are coherent and aligned with the complexity defined by the chosen category.

\paragraph{Final Query Synthesis}
In this phase, the structured contextual components are synthesized into a naturalistic user query. This is accomplished by leveraging the LLM as a "User Simulator", guided by a comprehensive prompt that orchestrates the generation process. The prompt instructs the LLM to generate an utterance that is authentic to the specified user's persona in communication style and vocabulary, and is also carefully constructed to be informationally self-contained for single-turn resolution. Furthermore, the query's structure must naturally imply the required tool interaction pattern (e.g., parallel or sequential calls) of its designated category. Crucially, the generated query remains strictly goal-oriented, focusing on the user's objective while avoiding any reference to the underlying tools or the process of task execution. The resulting output is a high-fidelity simulated user query, which then proceeds to the reference answer annotation stage.

\subsubsection{Reference Answer Annotation}
\label{sec:annotation}

The final stage in the construction of our benchmark is the annotation of a gold-standard reference answer for each query. Recognizing that the manual annotation of complex, multi-step agent tasks is both labor-intensive and requires significant domain expertise, we devised a hybrid, expert-in-the-loop framework. This methodology is engineered to maximize annotation quality and scalability by strategically directing human effort toward the most challenging cases, ensuring the overall robustness of the resulting benchmark.

Our annotation process unfolds in a two-tiered sequence, beginning with a broad automated assessment. First, for every query, an LLM is leveraged to generate $K$ distinct execution trajectories. These trajectories are then programmatically evaluated by a separate LLM-based judge to compute an aggregate pass rate. Queries that fall below a predetermined success threshold are automatically isolated and escalated for manual intervention. This initial screening efficiently filters out straightforward cases and pinpoints queries that are ambiguous, inherently difficult, or expose model limitations.

In the second stage, these flagged, low-pass-rate queries are subjected to a meticulous review by human experts. The expert's role extends beyond simple verification; they perform a diagnostic analysis to determine the root cause of failure. Based on their findings, the experts undertake one of several corrective actions: (1) revising the source query itself to resolve ambiguity or logical flaws, (2) refining a syntactically correct but semantically suboptimal LLM-generated answer, or (3) authoring a definitive reference answer from scratch. This targeted approach ensures that the most complex and nuanced instances within MCP-AgentBench are backed by human-validated, high-quality reference solutions. The final benchmark is thus composed of these curated query-answer pairs. We deliberately abstain from annotating execution trajectories as ground truth, as multiple valid solution paths often exist for a given task. For any queries involving time-sensitive information, the reference answer reflects a correct outcome as captured at the time of annotation.

\subsection{Data Statistics}
\label{sec:data_stats}

The resultant MCP-AgentBench dataset encompasses \textbf{600} unique queries, which are uniformly distributed with 100 queries designated for each of the six interaction categories defined in our methodology. This balanced distribution ensures comprehensive coverage of varied interaction complexities. Moreover, recognizing that the tool calling mode in most LLMs is typically restricted to a maximum of 128 tools, we randomized server selection within the \text{multi-server} categories to maintain the aggregate tool count near this practical threshold.

\subsection{MCP-Eval}
\label{sec:mcp_eval}
To facilitate a scalable and objective assessment of agent performance on MCP-AgentBench, we introduce MCP-Eval, an automated evaluation framework. MCP-Eval deliberately prioritizes the correctness of the final outcome over the intermediate execution trajectory, a design choice that acknowledges an agent's capacity for self-correction and the existence of multiple valid solution paths. The primary performance metric is the overall \textbf{Pass Rate}, which measures the fraction of queries an agent, $M_{eval}$, successfully resolves across the benchmark $\mathcal{B}$:
$$\text{Pass Rate} = \frac{\sum_{i=1}^{N} \mathbb{I}(J^{(i)} = \text{Pass})}{N}$$
where $N = |\mathcal{B}|$, $J^{(i)}$ is the binary judgment for test instance $i$, and $\mathbb{I}(\cdot)$ is the indicator function.

Our framework implements this outcome-oriented evaluation using an LLM-as-a-judge paradigm \cite{zheng2023judging}, employing a designated LLM, $L_{judge}$ (in our case, o3-mini-high \cite{OpenAI-o3-mini}). For each test instance, the judge assesses the model's generated answer $A_{model}^{(i)}$ against the user query $Q_{final}^{(i)}$ and the reference answer $A_{ref}^{(i)}$. This process is formalized as:
$$J^{(i)} = L_{judge}(Q_{final}^{(i)}, A_{ref}^{(i)}, A_{model}^{(i)}, \mathcal{I}_{judge})$$
The judge's decision is guided by a comprehensive prompt, $\mathcal{I}_{judge}$, which operationalizes a set of core principles to ensure objective and realistic assessment by prioritizing functional success over procedural mimicry. Key tenets include: \textbf{prioritizing the user's core intent} to define success; treating the \textbf{presence of specific, external data as conclusive evidence} of tool use, making its absence a primary failure criterion; emphasizing \textbf{sufficiency}, where meeting the core need is valued over exhaustive detail; and \textbf{disregarding superficial aspects} such as formatting or verbosity when comparing against the reference answer. 
By adhering to these instructions, $L_{judge}$ provides a scalable and consistent method for measuring practical agent effectiveness within the flexible and potentially non-deterministic context of complex tool interactions.

\section{Experiments}
\subsection{Experimental Setup} 

We conducted a comprehensive evaluation of 10 representative LLMs on MCP-AgentBench, encompassing both state-of-the-art proprietary systems and prominent open-source architectures. 
All models evaluated in this study were accessed uniformly through their respective official APIs.
These include proprietary systems such as Anthropic (\texttt{claude-3.7-sonnet, claude-4-sonnet}) \cite{Claude37}, OpenAI (\texttt{o3-mini-high, gpt-4o-2024-11-20}) \cite{achiam2023gpt}, and Google (\texttt{gemini-2.5-flash, gemini-2.5-pro}) \cite{comanici2025gemini}, alongside leading open-source models including Qwen (\texttt{qwen3-235b-a22b-thinking-2507}) \cite{qwen3technicalreport}, MoonshotAI (\texttt{kimi-k2}) \cite{team2025kimi}, and DeepSeek (\texttt{deepseek-v3-0324, deepseek-r1-0528}) \cite{deepseekai2024deepseekv3technicalreport, deepseekr1}. 

The ReAct framework served as the primary mechanism for enabling agent interaction. Additionally, native tool calling (TC) was employed for models possessing this inherent capability. However, Gemini's tool calling mode was not evaluated due to a formatting incompatibility between the tool descriptions on our deployed servers and the model's parser, which prevented successful execution. Agent activity for each query was constrained to a maximum of 30 actions, which could comprise either tool calls or textual responses. The interaction loop concludes and the agent formulates its final answer when a model output contains only text and no further tool calls, signaling that all necessary tool-assisted operations are considered complete. We evaluate the deep reasoning capabilities of advanced LLMs, such as Claude and Gemini, by configuring their "thinking budget" to 8192.

\subsection{Main Results}

\begin{table*}[!htbp]
\centering
\small
\begin{tabular}{ll l ccc ccc c} 
\toprule
\textbf{Type} & \textbf{Method} & \textbf{Model} & \multicolumn{3}{c}{\textbf{Single Server}} & \multicolumn{3}{c}{\textbf{Multi Server}} & \textbf{Avg.} \\
\cmidrule(lr){4-6} \cmidrule(lr){7-9}
& & & Single & Parallel & Sequential & Single & Parallel & Sequential & \\
\midrule
\multirow{10}{*}{\textbf{Proprietary}} & \multirow{6}{*}{ReAct} 
   & GPT-4o & 47 & 35 & 30 & 33 & 13 & 9 & 27.8 \\
&   & o3-mini & 56 & \textbf{63} & 41 & 48 & \textbf{50} & \textbf{49} & \textbf{51.2} \\
&   & Claude 3.7 Sonnet & 60 & 52 & 51 & 58 & 32 & 39 & 48.7 \\
&    & Claude 4 Sonnet & \textbf{62} & 50 & \textbf{57} & \textbf{66} & 27 & 33 & 49.2 \\
&   & Gemini 2.5 Flash & 54 & 54 & 42 & 63 & 30 & 34 & 46.2 \\
&   & Gemini 2.5 Pro & 58 & 53 & 43 & 58 & 38 & 44 & 49.0 \\
\cmidrule(l){2-10}
& \multirow{4}{*}{TC} 
    & GPT-4o & 53 & 41 & 25 & 38 & 12 & 15 & 30.7 \\
&   & o3-mini & \textbf{64} & 52 & 41 & 66 & 33 & 44 & 50.0 \\
&   & Claude 3.7 Sonnet & 63 & 48 & 47 & 69 & 39 & 52 & 53.0 \\
&    & Claude 4 Sonnet & 61 & \textbf{54} & \textbf{52} & \textbf{74} & \textbf{53} & \textbf{54} & \textbf{58.0} \\
\midrule
\multirow{7}{*}{\textbf{Open-source}} & \multirow{4}{*}{ReAct} 
   & Kimi K2  & 70 & 64 & \textbf{69} & 63 & 52 & 41 & 59.8 \\
&   & DeepSeek V3 & 59 & 58 & 55 & 61 & 36 & 45 & 52.3 \\
&    & DeepSeek R1 & 75 & 57 & 53 & 58 & 42 & 40 & 54.2 \\
&   & Qwen3-235B-A22B & \textbf{78} & \textbf{69} & \textbf{69} & \textbf{68} & \textbf{54} & \textbf{50} & \textbf{64.7} \\
\cmidrule(l){2-10}
& \multirow{3}{*}{TC} 
   & Kimi K2 & 63 & \textbf{61} & \textbf{65} & \textbf{70} & \textbf{48} & \textbf{59} & \textbf{61.0} \\
&    & DeepSeek V3 & 59 & 38 & 35 & 58 & 36 & 26 & 42.0 \\
&   & Qwen3-235B-A22B & \textbf{68} & 57 & \textbf{65} & 20 & 14 & 17 & 40.2 \\
\bottomrule
\end{tabular}
\caption{Main results of MCP-AgentBench (hierarchical structure). \textbf{Bold} denotes the best score within each sub-group. \textbf{Avg.} reflects the overall performance.}
\label{tab:main_results}
\end{table*}

\paragraph{Model Comparsion.}

The results in Table~\ref{tab:main_results} reveal a surprising and significant trend: the leading open-source models demonstrate exceptional capabilities, rivaling and even surpassing their proprietary counterparts. Most notably, Qwen3-235B-A22B, using the ReAct framework, achieved the highest overall score in the entire benchmark, challenging the prevailing narrative of proprietary model dominance. Among other open-source models, Kimi K2 also delivered robust results, particularly excelling in TC mode.
Within the proprietary category, Anthropic's Claude 4 Sonnet emerged as the top performer, showing a clear advantage when using its native TC capabilities. OpenAI's o3-mini also proved to be a consistent performer across both modes. In stark contrast, GPT-4o underperformed significantly in all tested scenarios, indicating potential limitations in its agentic reasoning for these tasks.

\paragraph{Method Comparsion.}
A comparison between the ReAct and TC modes highlights that model performance is highly dependent on the interaction framework, with no universally superior option. The most striking instance is Qwen3-235B-A22B, which leads the benchmark with ReAct but suffers a drastic performance collapse in TC mode. This failure often stems from the model not generating a tool call when one is required, leading to premature termination and an incorrect final answer. Conversely, models like Claude 4 Sonnet show a marked improvement with TC over ReAct, indicating its architecture is finely tuned for this interaction style. This variance underscores the critical importance of selecting a model-appropriate framework to unlock an agent's maximum potential.

\subsection{Analysis}
Subsequent analysis in this section will exclusively utilize results from the native TC mode for models supporting it. 

\paragraph{Demonstrating Varying Task Difficulty via Server Scope and Call Dependency}

\begin{figure}[htbp] 
\centering
\begin{minipage}{0.5\textwidth}
    \centering
    \includegraphics[width=\linewidth]{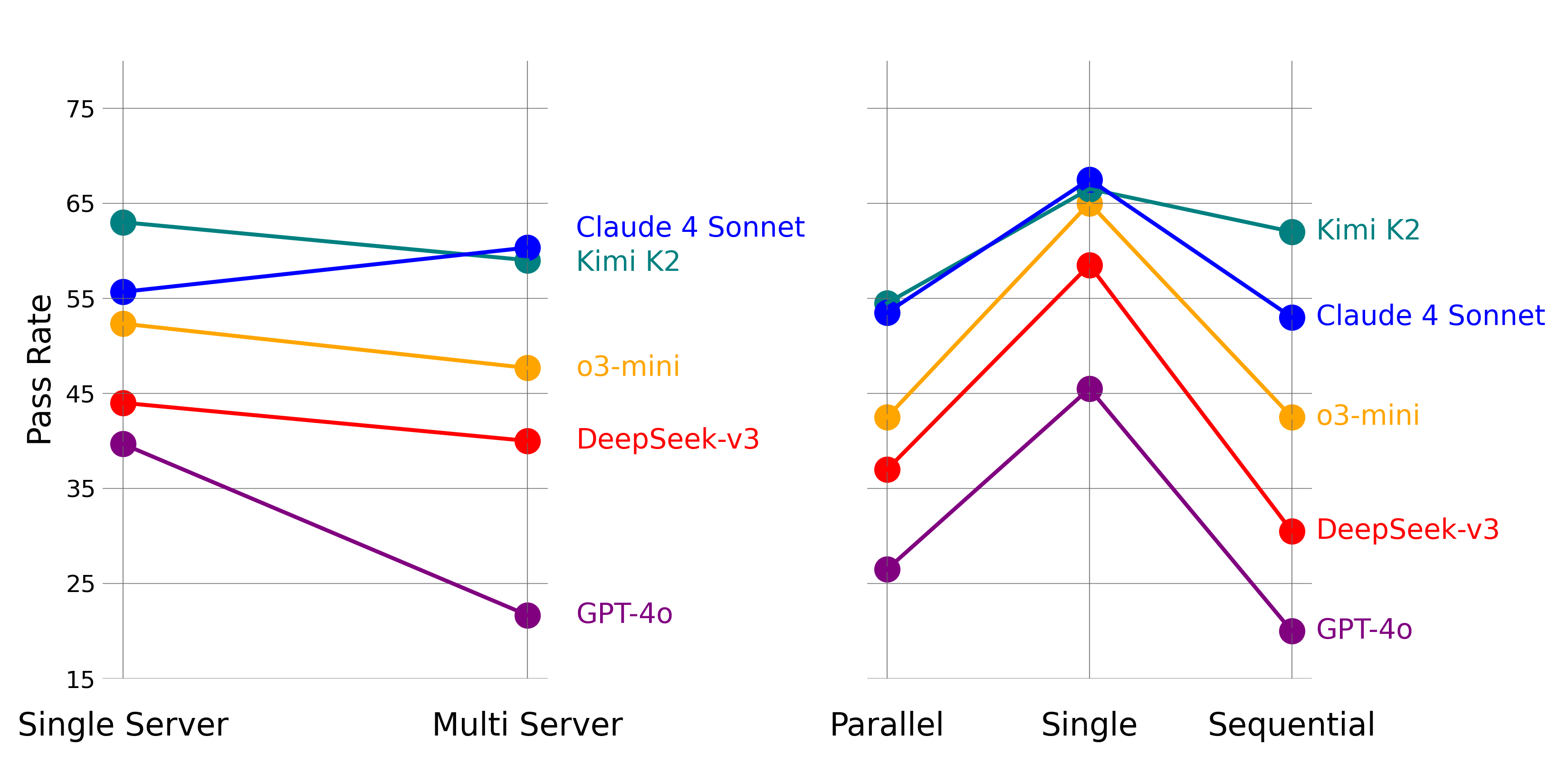}
\end{minipage}
\hfill 
\begin{minipage}{0.48\textwidth}
    \centering
    \includegraphics[width=\linewidth]{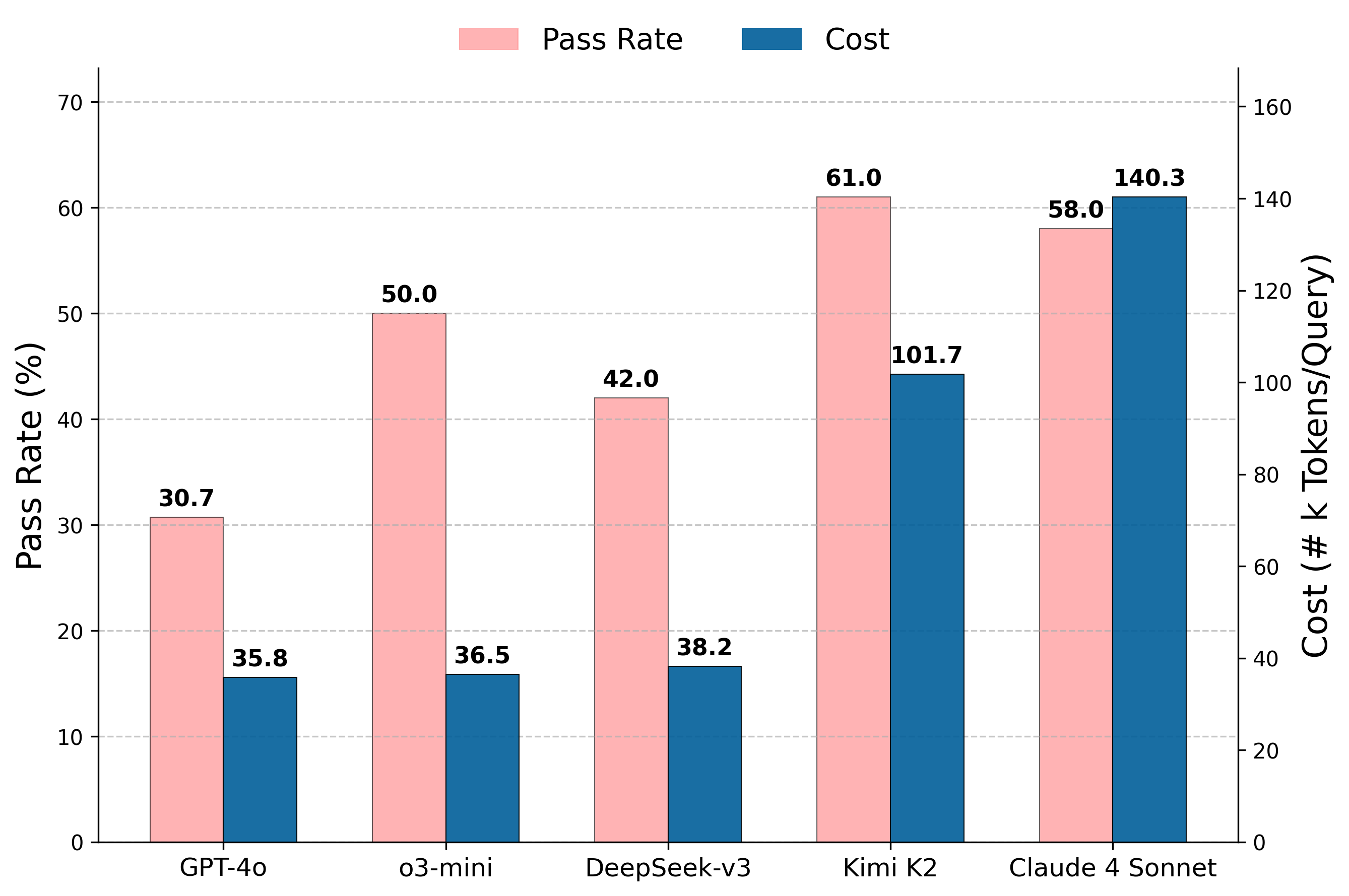}
    \vspace{-0.5em}
\end{minipage}

\vspace{0.5em} 

\begin{minipage}{0.48\textwidth}
    \centering
    \caption{Model pass rates under different server scopes (left) and call dependencies (right), illustrating varying task difficulty.}
    \label{fig:perf_server_scope_and_dependency}
\end{minipage}
\hfill
\begin{minipage}{0.46\textwidth}
    \centering
    \caption{Model performance vs. token consumption.}
    \label{fig:cost} 
\end{minipage}
\end{figure}

MCP-AgentBench's graduated difficulty is evident across two dimensions. First, performance generally declines as tasks transition from Single Server to Multi Server scopes. Second, a similar drop occurs as call dependency increases from simple Single to complex Sequential calls. Claude 4 Sonnet, however, is a notable exception to the first trend, showing an improved pass rate on more challenging multi-server tasks. Our analysis indicates this anomaly occurs because the model's tendency to fail on simpler tasks by incorrectly relying on its parametric knowledge is mitigated when greater complexity compels reliable tool engagement. This hierarchy of challenges confirms MCP-AgentBench's ability to test a wide spectrum of agentic capabilities, as shown in Figure~\ref{fig:perf_server_scope_and_dependency}.

\paragraph{Token Efficiency.}

Figure~\ref{fig:cost} illustrates the trade-off between model performance and token consumption. It reveals that the models with the highest pass rates, Kimi K2 (61.0\%) and Claude 4 Sonnet (58.0\%), are also by far the most costly, consuming 101.7k and 140.3k tokens per query, respectively. This higher consumption is largely attributable to their use of a "thinking" mode for internal reasoning before generating a response. In contrast, o3-mini demonstrates exceptional efficiency. It achieves a strong 50.0\% pass rate with a token cost of only 36.5k, which is comparable to models with significantly lower performance, such as GPT-4o and DeepSeek-v3 . This disparity highlights that while higher performance often correlates with increased token usage, some models like o3-mini offer a significantly more cost-effective balance.

\paragraph{Consistency between MCP-Eval and Human Evaluation}
We assessed the consistency between MCP-Eval outputs and human evaluations, alongside inter-rater reliability. Three human experts annotated 60 items, comprising 10 randomly selected items from each of six categories derived from Claude 3.7 Sonnet outputs. This annotation task required each expert to dedicate approximately 2.5 hours (150 minutes) in total, averaging 2.5 minutes per item. The percentage agreement between MCP-Eval and the aggregated human majority vote reached 91.67\%, with a Cohen's Kappa of 0.734. For inter-rater reliability on pass/fail decisions, Fleiss' Kappa among the experts was 0.671. The overall three-way percentage agreement for all valid expert judgments stood at 86.67\%. This strong concordance validates MCP-Eval's alignment with human evaluative standards and confirms the robustness of our human-annotated dataset. The calculation methods for the evaluation metrics discussed above are detailed in Appendix~\ref{consistency}.

\paragraph{Error Analysis}
Based on our evaluation of agent performance on MCP-AgentBench, we identified several recurring categories of errors. This analysis provides insight into the common failure modes of LLM agents in protocol-driven, tool-use scenarios.

\begin{itemize}
    \item \textbf{Misinterpretation of Query}:
    The agent fails to accurately capture the user's primary objective or misconstrues critical semantic details and constraints embedded within the query.

    \item \textbf{Refusal to Use Tool}:
    The agent improperly defaults to its parametric knowledge, providing a response from its internal memory when the task explicitly necessitates tool invocation for accessing external, dynamic, or proprietary data sources.

    \item \textbf{Omission of Key Information}:
    The agent delivers an incomplete response, either by failing to address all components of the user's request or by neglecting to synthesize and incorporate essential information obtained from tool outputs in previous steps of a multi-turn interaction.

    \item \textbf{Hallucination}:
    The agent fabricates information, presenting factually incorrect details or assertions that are either unsupported by or directly contradict the evidence provided by the tool outputs.
\end{itemize}

\section{Related Work}

\paragraph{LLM Agents and Standardized Interaction}
The proliferation of sophisticated LLM-based agents, capable of complex reasoning and planning \cite{huang2022towards, huang2024understanding}, has unlocked new capabilities in domains like web navigation, software engineering, and GUI automation \cite{wang2024survey, guo2024large, zhang2024large}. A key enabler for these agents is their ability to interact with external tools \cite{patil2024gorilla, qin2023toolllm}, a process facilitated by frameworks like AutoGen \cite{wu2023autogenenablingnextgenllm} and MetaGPT \cite{hong2024metagpt}. However, scaling these interactions presents a significant "M$\times$N integration problem." In response, standardized communication frameworks like the Model Context Protocol (MCP) \cite{mcp} have been developed to provide a universal layer between agents and tools. The adoption of such protocols necessitates new benchmarks specifically designed to evaluate agent performance within these structured environments.

\paragraph{Benchmarking Protocol-Aware Tool Use}
Existing tool-use benchmarks, such as ToolBench \cite{qin2023toolllm} and API-Bank \cite{li2023apibankcomprehensivebenchmarktoolaugmented}, have been instrumental in assessing general agent capabilities. Yet, they were not designed to measure agent effectiveness within the context of a standardized protocol, a critical aspect for real-world deployment. While evaluations for the MCP ecosystem are emerging, they have key limitations. MCP-Bench \cite{mcpbench_report}, for instance, evaluates the performance of MCP servers, not the proficiency of agents. Others, such as MCPEval \cite{liu2025mcpeval} and MCP-RADAR \cite{gao2025mcp}, are constrained by a limited scope of servers and specific domains, which compromises their comprehensiveness. This leaves a critical gap for a large-scale benchmark that assesses agents across a diverse and operational MCP infrastructure. To address this, we introduce MCP-AgentBench, a comprehensive benchmark designed to rigorously evaluate agent performance in such environments.

\section{Conclusion}
This work addresses the critical evaluation gap for language agents within the Model Context Protocol (MCP) ecosystem by delivering three key contributions. First, we established the MCP Server Testbed, a large-scale operational infrastructure (33 servers, 188 tools) that provides a realistic, standardized environment. Second, we built MCP-AgentBench, a comprehensive benchmark featuring 600 diverse queries designed to rigorously assess agent performance on this testbed. Third, we introduced MCP-Eval, a novel, outcome-oriented LLM-as-a-judge methodology that prioritizes tangible task success. Collectively, our work provides a standardized framework and crucial empirical insights to accelerate the development of capable and truly interoperable AI systems that fully leverage the MCP standard.

\bibliographystyle{plain}
\bibliography{reference}

\begin{thebibliography}{10}

\bibitem{achiam2023gpt}
Josh Achiam, Steven Adler, Sandhini Agarwal, Lama Ahmad, Ilge Akkaya,
  Florencia~Leoni Aleman, Diogo Almeida, Janko Altenschmidt, Sam Altman,
  Shyamal Anadkat, et~al.
\newblock Gpt-4 technical report.
\newblock {\em arXiv preprint arXiv:2303.08774}, 2023.

\bibitem{Claude-computeruse}
Anthropic.
\newblock Introducing computer use, a new claude 3.5 sonnet, and claude 3.5
  haiku.
\newblock \url{https://www.anthropic.com/news/3-5-models-and-computer-use},
  2024.

\bibitem{Claude37}
Anthropic.
\newblock Claude 3.7 sonnet \ anthropic.
\newblock \url{https://www.anthropic.com/claude/sonnet}, 2025.

\bibitem{mcp}
Anthropic.
\newblock Introducing the model context protocol \ anthropic.
\newblock \url{https://www.anthropic.com/news/model-context-protocol}, 2025.

\bibitem{chatmcpmcprouter}
chatmcp.
\newblock chatmcp/mcprouter: api router for mcp servers.
\newblock \url{https://github.com/chatmcp/mcprouter}, 2025.

\bibitem{comanici2025gemini}
Gheorghe Comanici, Eric Bieber, Mike Schaekermann, Ice Pasupat, Noveen
  Sachdeva, Inderjit Dhillon, Marcel Blistein, Ori Ram, Dan Zhang, Evan Rosen,
  et~al.
\newblock Gemini 2.5: Pushing the frontier with advanced reasoning,
  multimodality, long context, and next generation agentic capabilities.
\newblock {\em arXiv preprint arXiv:2507.06261}, 2025.

\bibitem{deepseekai2024deepseekv3technicalreport}
DeepSeek-AI.
\newblock Deepseek-v3 technical report, 2024.

\bibitem{deepseekr1}
DeepSeek-AI.
\newblock Deepseek-r1: Incentivizing reasoning capability in llms via
  reinforcement learning, 2025.

\bibitem{gao2025mcp}
Xuanqi Gao, Siyi Xie, Juan Zhai, Shqing Ma, and Chao Shen.
\newblock Mcp-radar: A multi-dimensional benchmark for evaluating tool use
  capabilities in large language models.
\newblock {\em arXiv preprint arXiv:2505.16700}, 2025.

\bibitem{guo2024large}
Taicheng Guo, Xiuying Chen, Yaqi Wang, Ruidi Chang, Shichao Pei, Nitesh~V
  Chawla, Olaf Wiest, and Xiangliang Zhang.
\newblock Large language model based multi-agents: A survey of progress and
  challenges.
\newblock {\em arXiv preprint arXiv:2402.01680}, 2024.

\bibitem{hong2024metagpt}
Sirui Hong, Mingchen Zhuge, Jonathan Chen, Xiawu Zheng, Yuheng Cheng, Jinlin
  Wang, Ceyao Zhang, Zili Wang, Steven Ka~Shing Yau, Zijuan Lin, Liyang Zhou,
  Chenyu Ran, Lingfeng Xiao, Chenglin Wu, and J{\"u}rgen Schmidhuber.
\newblock Meta{GPT}: Meta programming for a multi-agent collaborative
  framework.
\newblock In {\em The Twelfth International Conference on Learning
  Representations}, 2024.

\bibitem{hou2025model}
Xinyi Hou, Yanjie Zhao, Shenao Wang, and Haoyu Wang.
\newblock Model context protocol (mcp): Landscape, security threats, and future
  research directions.
\newblock {\em arXiv preprint arXiv:2503.23278}, 2025.

\bibitem{huang2022towards}
Jie Huang and Kevin Chen-Chuan Chang.
\newblock Towards reasoning in large language models: A survey.
\newblock {\em arXiv preprint arXiv:2212.10403}, 2022.

\bibitem{huang2024understanding}
Xu~Huang, Weiwen Liu, Xiaolong Chen, Xingmei Wang, Hao Wang, Defu Lian, Yasheng
  Wang, Ruiming Tang, and Enhong Chen.
\newblock Understanding the planning of llm agents: A survey.
\newblock {\em arXiv preprint arXiv:2402.02716}, 2024.

\bibitem{li2023apibankcomprehensivebenchmarktoolaugmented}
Minghao Li, Yingxiu Zhao, Bowen Yu, Feifan Song, Hangyu Li, Haiyang Yu, Zhoujun
  Li, Fei Huang, and Yongbin Li.
\newblock Api-bank: A comprehensive benchmark for tool-augmented llms, 2023.

\bibitem{liu2024large}
Junwei Liu, Kaixin Wang, Yixuan Chen, Xin Peng, Zhenpeng Chen, Lingming Zhang,
  and Yiling Lou.
\newblock Large language model-based agents for software engineering: A survey.
\newblock {\em arXiv preprint arXiv:2409.02977}, 2024.

\bibitem{liu2025mcpeval}
Zhiwei Liu, Jielin Qiu, Shiyu Wang, Jianguo Zhang, Zuxin Liu, Roshan Ram,
  Haolin Chen, Weiran Yao, Huan Wang, Shelby Heinecke, et~al.
\newblock Mcpeval: Automatic mcp-based deep evaluation for ai agent models.
\newblock {\em arXiv preprint arXiv:2507.12806}, 2025.

\bibitem{mcpbench_report}
Zhiling Luo, Xiaorong Shi, Xuanrui Lin, and Jinyang Gao.
\newblock Evaluation report on mcp servers.
\newblock {\em arXiv preprint arXiv:2504.11094}, 2025.

\bibitem{niu2024screenagent}
Runliang Niu, Jindong Li, Shiqi Wang, Yali Fu, Xiyu Hu, Xueyuan Leng, He~Kong,
  Yi~Chang, and Qi~Wang.
\newblock Screenagent: A vision language model-driven computer control agent.
\newblock {\em arXiv preprint arXiv:2402.07945}, 2024.

\bibitem{gpt4o}
OpenAI.
\newblock Hello gpt-4o | openai.
\newblock \url{https://openai.com/index/hello-gpt-4o/}, 2025.

\bibitem{OpenAI-o3-mini}
OpenAI.
\newblock Openai o3-mini | openai.
\newblock \url{https://openai.com/index/openai-o3-mini/}, 2025.

\bibitem{openrouter}
OpenRouter.
\newblock Llm rankings | openrouter.
\newblock \url{https://openrouter.ai/rankings?view=month}, 2025.

\bibitem{pan2024webcanvas}
Yichen Pan, Dehan Kong, Sida Zhou, Cheng Cui, Yifei Leng, Bing Jiang, Hangyu
  Liu, Yanyi Shang, Shuyan Zhou, Tongshuang Wu, et~al.
\newblock Webcanvas: Benchmarking web agents in online environments.
\newblock {\em arXiv preprint arXiv:2406.12373}, 2024.

\bibitem{patil2024gorilla}
Shishir~G Patil, Tianjun Zhang, Xin Wang, and Joseph~E Gonzalez.
\newblock Gorilla: Large language model connected with massive apis.
\newblock {\em Advances in Neural Information Processing Systems},
  37:126544--126565, 2024.

\bibitem{qin2023toolllm}
Yujia Qin, Shihao Liang, Yining Ye, Kunlun Zhu, Lan Yan, Yaxi Lu, Yankai Lin,
  Xin Cong, Xiangru Tang, Bill Qian, Sihan Zhao, Runchu Tian, Ruobing Xie, Jie
  Zhou, Mark Gerstein, Dahai Li, Zhiyuan Liu, and Maosong Sun.
\newblock Toolllm: Facilitating large language models to master 16000+
  real-world apis, 2023.

\bibitem{su2024language}
Yu~Su, Diyi Yang, Shunyu Yao, and Tao Yu.
\newblock Language agents: Foundations, prospects, and risks.
\newblock In {\em Proceedings of the 2024 Conference on Empirical Methods in
  Natural Language Processing: Tutorial Abstracts}, pages 17--24, 2024.

\bibitem{sumers2023cognitive}
Theodore Sumers, Shunyu Yao, Karthik Narasimhan, and Thomas Griffiths.
\newblock Cognitive architectures for language agents.
\newblock {\em Transactions on Machine Learning Research}, 2023.

\bibitem{team2025kimi}
Kimi Team, Yifan Bai, Yiping Bao, Guanduo Chen, Jiahao Chen, Ningxin Chen,
  Ruijue Chen, Yanru Chen, Yuankun Chen, Yutian Chen, et~al.
\newblock Kimi k2: Open agentic intelligence.
\newblock {\em arXiv preprint arXiv:2507.20534}, 2025.

\bibitem{qwen3technicalreport}
Qwen Team.
\newblock Qwen3 technical report, 2025.

\bibitem{wang2024survey}
Lei Wang, Chen Ma, Xueyang Feng, Zeyu Zhang, Hao Yang, Jingsen Zhang, Zhiyuan
  Chen, Jiakai Tang, Xu~Chen, Yankai Lin, et~al.
\newblock A survey on large language model based autonomous agents.
\newblock {\em Frontiers of Computer Science}, 18(6):186345, 2024.

\bibitem{wu2023autogenenablingnextgenllm}
Qingyun Wu, Gagan Bansal, Jieyu Zhang, Yiran Wu, Beibin Li, Erkang Zhu,
  Li~Jiang, Xiaoyun Zhang, Shaokun Zhang, Jiale Liu, Ahmed~Hassan Awadallah,
  Ryen~W White, Doug Burger, and Chi Wang.
\newblock Autogen: Enabling next-gen llm applications via multi-agent
  conversation, 2023.

\bibitem{xie2024osworld}
Tianbao Xie, Danyang Zhang, Jixuan Chen, Xiaochuan Li, Siheng Zhao, Ruisheng
  Cao, Toh~J Hua, Zhoujun Cheng, Dongchan Shin, Fangyu Lei, et~al.
\newblock Osworld: Benchmarking multimodal agents for open-ended tasks in real
  computer environments.
\newblock {\em Advances in Neural Information Processing Systems},
  37:52040--52094, 2024.

\bibitem{xue2025illusion}
Tianci Xue, Weijian Qi, Tianneng Shi, Chan~Hee Song, Boyu Gou, Dawn Song, Huan
  Sun, and Yu~Su.
\newblock An illusion of progress? assessing the current state of web agents.
\newblock {\em arXiv preprint arXiv:2504.01382}, 2025.

\bibitem{berkeley-function-calling-leaderboard}
Fanjia Yan, Huanzhi Mao, Charlie Cheng-Jie Ji, Tianjun Zhang, Shishir~G. Patil,
  Ion Stoica, and Joseph~E. Gonzalez.
\newblock Berkeley function calling leaderboard.
\newblock
  \url{https://gorilla.cs.berkeley.edu/blogs/8_berkeley_function_calling_leaderboard.html},
  2024.

\bibitem{yang2024swe}
John Yang, Carlos Jimenez, Alexander Wettig, Kilian Lieret, Shunyu Yao, Karthik
  Narasimhan, and Ofir Press.
\newblock Swe-agent: Agent-computer interfaces enable automated software
  engineering.
\newblock {\em Advances in Neural Information Processing Systems},
  37:50528--50652, 2024.

\bibitem{yao2024tau}
Shunyu Yao, Noah Shinn, Pedram Razavi, and Karthik Narasimhan.
\newblock $\tau$-bench: A benchmark for tool-agent-user interaction in
  real-world domains.
\newblock {\em arXiv preprint arXiv:2406.12045}, 2024.

\bibitem{zhang2024large}
Chaoyun Zhang, Shilin He, Jiaxu Qian, Bowen Li, Liqun Li, Si~Qin, Yu~Kang,
  Minghua Ma, Guyue Liu, Qingwei Lin, et~al.
\newblock Large language model-brained gui agents: A survey.
\newblock {\em arXiv preprint arXiv:2411.18279}, 2024.

\bibitem{zheng2023judging}
Lianmin Zheng, Wei-Lin Chiang, Ying Sheng, Siyuan Zhuang, Zhanghao Wu, Yonghao
  Zhuang, Zi~Lin, Zhuohan Li, Dacheng Li, Eric Xing, et~al.
\newblock Judging llm-as-a-judge with mt-bench and chatbot arena.
\newblock {\em Advances in Neural Information Processing Systems},
  36:46595--46623, 2023.

\bibitem{zhong2025complexfuncbench}
Lucen Zhong, Zhengxiao Du, Xiaohan Zhang, Haiyi Hu, and Jie Tang.
\newblock Complexfuncbench: Exploring multi-step and constrained function
  calling under long-context scenario, 2025.

\bibitem{zhou2023webarena}
Shuyan Zhou, Frank~F Xu, Hao Zhu, Xuhui Zhou, Robert Lo, Abishek Sridhar,
  Xianyi Cheng, Tianyue Ou, Yonatan Bisk, Daniel Fried, et~al.
\newblock Webarena: A realistic web environment for building autonomous agents.
\newblock {\em arXiv preprint arXiv:2307.13854}, 2023.

\end{thebibliography}


\newpage
\appendix
\label{sec:appendix}


\section{Data Statistics}
\label{data}

\begin{figure}[ht]
  \centering 
  \begin{subfigure}[b]{0.48\textwidth} 
    \centering
    \includegraphics[width=\linewidth]{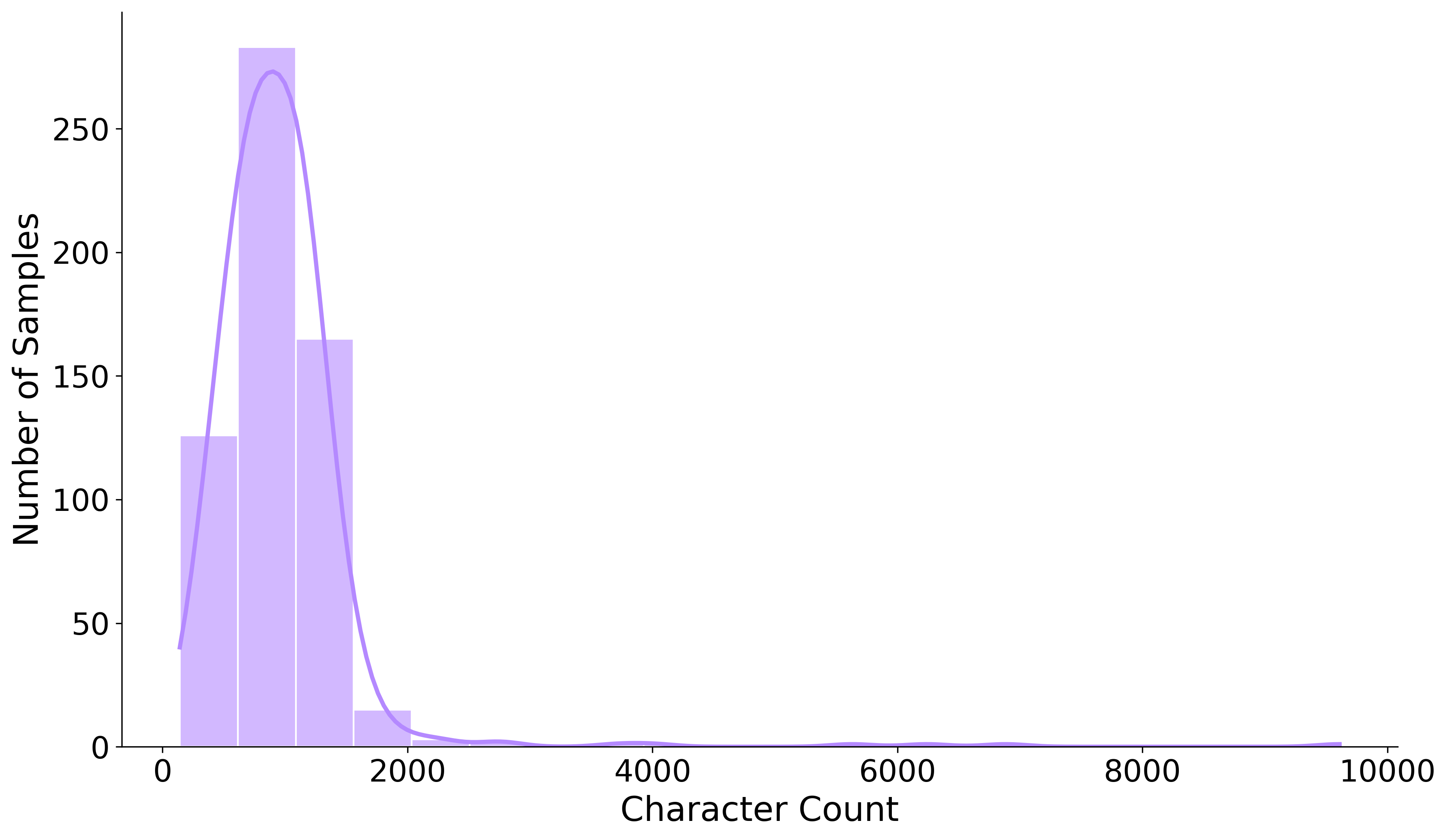}
    \caption{Query Length Distribution}
    \label{fig:query_length_dist}
  \end{subfigure}
  \hfill 
  \begin{subfigure}[b]{0.48\textwidth}
    \centering
    \includegraphics[width=\linewidth]{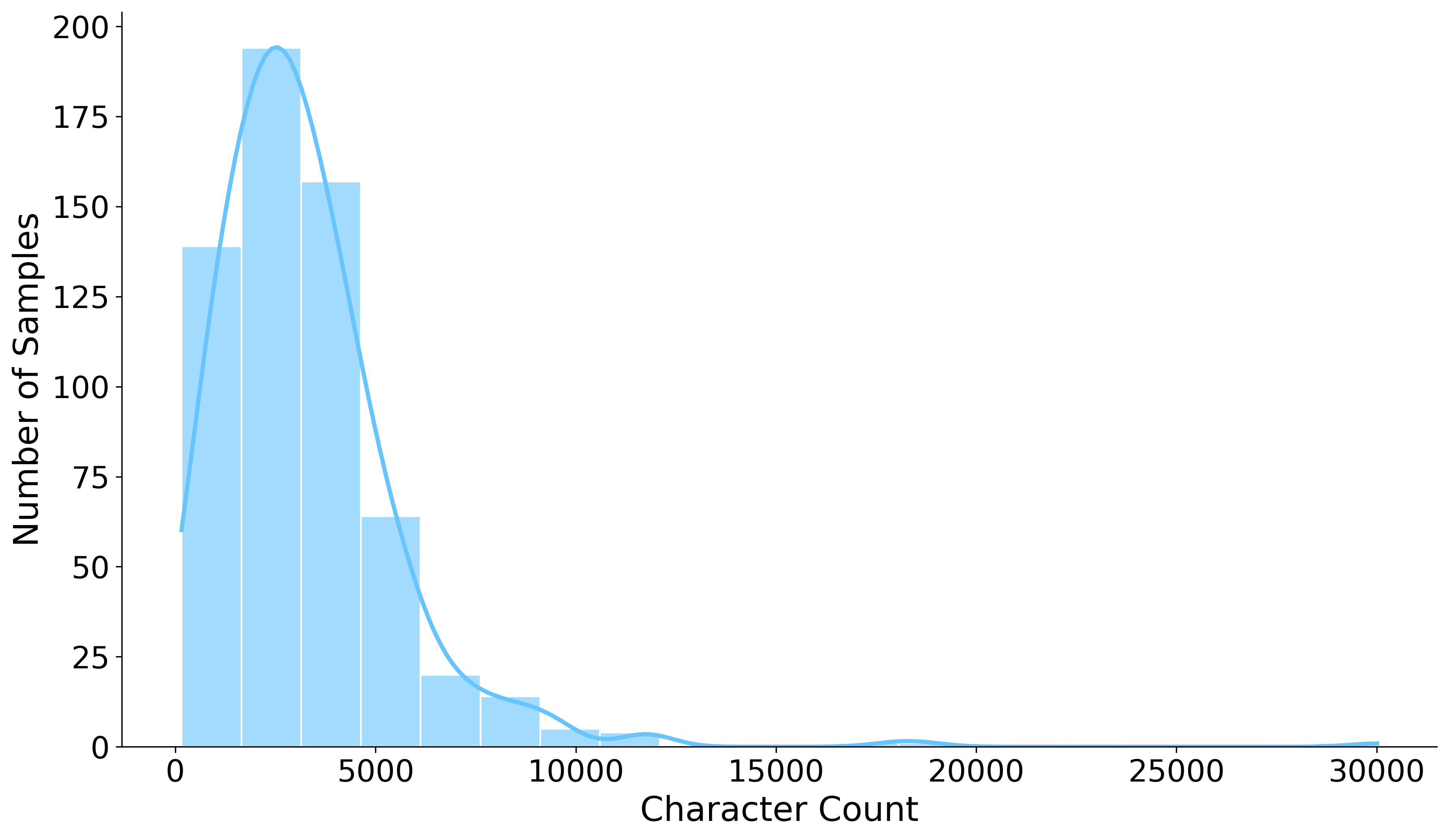}
    \caption{Reference Answer Length Distribution}
    \label{fig:answer_length_dist}
  \end{subfigure}

  \caption{Distribution of query and reference answer lengths of MCP-AgentBench.}
  \label{fig:length_distributions_combined}
\end{figure}

The analysis of query and answer length distributions reveals distinct textual characteristics and asymmetries within MCP-AgentBench. Input query lengths (Figure~\ref{fig:query_length_dist}) exhibit a right-skewed distribution, primarily peaking in the 800-1,000 character range; these lengths suggest moderately detailed input prompts. In contrast, reference answer lengths (Figure~\ref{fig:answer_length_dist}), while also right-skewed, are considerably longer and demonstrate greater variability, featuring modal values prominently within the 2,000-4,000 character range and a more pronounced tail. These distributions highlight a notable asymmetry between query and response: agents are expected to interpret inputs of moderate detail and subsequently generate substantially more extensive and varied outputs to successfully accomplish task objectives.

\section{Reference Answer Annotation Details}
In the reference answer annotation process described in the main paper, we specified that an LLM generates $K$ distinct execution trajectories for automated assessment. For our experiments, the value of $K$ was set to \textbf{5}. Queries were identified as "low-pass-rate" and escalated for manual expert review if their aggregate pass rate fell below a threshold of \textbf{20\%}.

\section{Experimental Details}
\label{experiment_detail}

\subsection{ReAct Format}
\label{react_format}

In our ReAct implementation, the \texttt{history} string sequentially captures the agent's interaction trajectory. It commences with the \textbf{User Query}. Subsequent agent activity unfolds in cycles, each typically documented by appending the agent's \textbf{Thought}, the \textbf{Action} taken (if any, e.g., tool invocation specifics), and the resulting \textbf{Observation} (tool execution outcome). Observations are distinctly formatted (e.g., with extra newlines) to clearly demarcate separate interaction cycles. The entire process concludes with the agent's \textbf{Final Answer}.

If a predefined maximum step limit is exceeded, a system directive is issued, prompting an immediate \textbf{Final Answer}. This structured \texttt{history} string, with its ordered entries, constitutes the complete contextual input for subsequent LLM prompts.

\subsection{Tool Calling Mode}
In Tool Calling mode, the interaction process is initiated when the model receives the user query alongside a list of available tools. The model then typically responds by issuing a specific tool call, detailing the selected tool and its input parameters. We have observed, for example, that while reasoning models like \texttt{o3-mini} and \texttt{Claude 3.7 Sonnet} tend to generate only a single tool call per step, other models such as \texttt{GPT-4o} are capable of generating multiple tool calls concurrently. Following the external execution of this designated tool, its resulting output (the observation) is provided back to the model. This cycle can then repeat, with the model making subsequent tool calls based on the updated conversational context and the outcomes of previous actions. The interaction trajectory is considered complete once the model generates a final answer without initiating any further tool calls.

\subsection{Configuration}
\begin{itemize}
    \item \textbf{Thinking Budget:} For models supporting enhanced reasoning capabilities, such as \texttt{claude-3.7-sonnet}, \texttt{gemini-2.5-pro}, and \texttt{gemini-2.5-flash}, we utilized their thinking modes with the thinking budget parameter set to 8192 tokens to maximize their analytical performance on this benchmark.
    \item \textbf{Output Length:} The maximum output length for model generations was set to 8192 tokens to ensure responses could be comprehensive and avoid premature truncation.
    \item \textbf{API Request Retries:} For all model API interactions, requests were retried up to 3 times in the event of a transient failure to ensure robustness.
\end{itemize}

\section{Calculation of Metrics for Consistency Evaluation}
\label{consistency} 

This appendix outlines the calculation of metrics used to evaluate MCP-Eval's consistency with human judgments and inter-rater reliability. These metrics were applied to $N=60$ items, classified into $k=2$ categories (e.g., "pass"/"fail") by MCP-Eval and $n=3$ human experts.

\subsection{Percentage Agreement (MCP-Eval vs. Human Majority)}
Measures the proportion of $N$ items where MCP-Eval's decision matches the human expert majority vote ($N_{\text{agree}}$):
$$P_o = \frac{N_{\text{agree}}}{N}$$

\subsection{Cohen's Kappa (MCP-Eval vs. Human Majority)}

Measures agreement between two raters (MCP-Eval and the human majority vote), correcting for chance agreement.
\begin{equation*}
    \kappa = \frac{P_o - P_e}{1 - P_e}
\end{equation*}
where:
\begin{itemize}
    \item $P_o$ is the observed proportional agreement (see above).
    \item $P_e$ is the probability of chance agreement. For categories $c_1, \dots, c_k$ (here $k=2$):
    $P_e = \sum_{j=1}^{k} P(\text{MCP-Eval assigns } c_j) \times P(\text{Human Majority assigns } c_j)$.
    Each $P(\text{Rater assigns } c_j)$ is the proportion of items assigned to category $c_j$ by that rater.
\end{itemize}

\subsection{Cohen's Kappa (MCP-Eval vs. Human Majority)}

Assesses agreement reliability among $n=3$ human experts for $N$ items and $k=2$ categories.
\begin{equation*}
    \kappa = \frac{\bar{P} - \bar{P}_e}{1 - \bar{P}_e}
\end{equation*}
where:
\begin{itemize}
    \item $\bar{P}$ is the mean observed proportion of rater agreement per item. For item $i$, let $n_{ij}$ be the number of raters assigning it to category $j$. Then the agreement for item $i$ is $P_i = \frac{1}{n(n-1)} \sum_{j=1}^{k} n_{ij}(n_{ij}-1)$, and $\bar{P} = \frac{1}{N} \sum_{i=1}^{N} P_i$.
    \item $\bar{P}_e$ is the expected chance agreement. Let $p_j = \frac{1}{Nn} \sum_{i=1}^{N} n_{ij}$ be the proportion of all assignments to category $j$. Then $\bar{P}_e = \sum_{j=1}^{k} p_j^2$.
\end{itemize}

\subsection{Overall Three-Way Percentage Agreement (Among Experts)}
Measures the proportion of $N$ items for which all $n=3$ human experts provided the identical rating ($N_{\text{all\_agree}}$):
$$P_{\text{3-way}} = \frac{N_{\text{all\_agree}}}{N}$$

\section{Prompts}
\label{sec:prompts}

\begin{tcolorbox}[
    enhanced,
    breakable,
    colback=blue!5!white,
    colframe=blue!75!black,
    title=Query Synthesis System Prompt,
    fonttitle=\bfseries
]

You are a User Simulator whose purpose is to generate authentic user queries that naturally express how a described person would communicate with an AI assistant. Your task is to create realistic, in-character messages that will lead to deterministic, verifiable results without being artificially explicit.

\textbf{CORE REQUIREMENTS}
\begin{itemize}
    \item Generate exactly what the person described in the <user\_profile> would say when trying to accomplish their <objective> within the given <scenario>
    \item The generated query MUST be perfectly solvable by the <selected\_tools> - this is a critical requirement
    \item This is a one-turn interaction - create a SINGLE message that includes the user's complete request
    \item Include all information necessary for the task to be completed without follow-up questions
    \item Structure the request to naturally follow the specified category pattern without explicitly mentioning the pattern
\end{itemize}

\textbf{SCENARIO CATEGORIES AND TOOL USAGE}

Each query must align precisely with one of these interaction patterns:
\begin{itemize}
    \item single\_server\_single\_call: Using 1 tool from 1 server with a single call
    \item single\_server\_parallel\_call: Using multiple tools or multiple calls to the same tool from 1 server with no dependencies between calls
    \item single\_server\_sequential\_call: Using multiple tools from 1 server with dependencies between calls, requiring specific order
    \item multi\_server\_single\_call: Selecting ONLY 1 tool from across multiple available servers and making a single call (not selecting multiple tools)
    \item multi\_server\_parallel\_call: Using multiple tools from multiple servers with no dependencies between calls
    \item multi\_server\_sequential\_call: Using multiple tools from multiple servers with dependencies between calls, requiring specific order
\end{itemize}

For sequential tasks, ensure the user's request implies the necessary dependencies that would require steps to be performed in order.

\textbf{CHARACTER AUTHENTICITY}

Ensure the message is completely in-character, reflecting:
\begin{itemize}
    \item Technical vocabulary and domain expertise appropriate to the user profile
    \item Communication style (formal/informal, verbose/concise)
    \item Emotional state given the scenario's pressures
    \item Typical sentence structure and word choice
    \item Natural expression of needs without artificial formatting
    \item Appropriate level of precision (e.g., a software engineer might use more precise terminology than a casual user)
\end{itemize}

\textbf{INFORMATION COMPLETENESS}

All information needed to complete the user's request can come from:

\begin{enumerate}
    \item Information naturally expressed in the user's initial request
    \item Reasonably inferred from context provided in the initial request
    \item Common knowledge that would be expected for the task
    \item Derived from previous tool results within the scenario (for multi-step tasks)
\end{enumerate}

When providing information:
\begin{itemize}
    \item Express information as this specific user naturally would, not in technical formats
    \item Use everyday language for entities, locations, times and dates
    \item Provide sufficient context for all necessary information to be unambiguously determined
    
\end{itemize}

\textbf{TIME SENSITIVITY CONSIDERATIONS}

When temporal information is relevant:
\begin{itemize}
    \item Include specific temporal context (date, day of week, time of day) as needed
    \item Express time-related information naturally as the user would (e.g., "this weekend" vs. "April 24-25, 2025")
    \item For urgent requests, reflect appropriate time pressure in the user's language
    \item Reference seasonal factors, holidays, or events when relevant
    \item Include deadlines or time constraints if they're important to the task
    \item Ensure time references are unambiguous with sufficient context
    \item For scheduling tasks, include clear temporal boundaries (start/end times, duration)
    \item For recurring events, specify the frequency pattern naturally
\end{itemize}

\textbf{AUTHENTICITY GUIDELINES AND LIMITATIONS}
\begin{itemize}
    \item DO NOT mention AI tools or suggest which tools should be used
    \item DO NOT include unnecessary self-introductions unless natural for this character
    \item DO NOT explicitly explain what would constitute success
    \item DO NOT include any meta-information about the user or simulation process
    \item Focus entirely on WHAT the user needs, not HOW the assistant should accomplish it
    \item Include context and constraints in a natural, conversational way
    \item Avoid requesting features or information that would require tools beyond those provided
\end{itemize}

\textbf{OUTPUT FORMAT}

Write only what the user would actually say or type, with no explanations, instructions, or meta-commentary. Your output should be a natural language query that:
\begin{enumerate}
    \item Covers all aspects of the user's objective
    \item Matches how this specific person would naturally communicate
    \item Contains sufficient information for deterministic task completion
    \item Remains authentic and conversational rather than structured as a technical request
    \item Naturally leads to the tool usage pattern specified by the selected category
    \item Can be completely and effectively addressed using only the <selected\_tools>
    \item Includes appropriate temporal context when the query is time-sensitive
\end{enumerate}
\end{tcolorbox}

\begin{tcolorbox}[
    breakable,
    colback=blue!5!white,
    colframe=blue!75!black,
    title=Query Synthesis User Prompt,
    fonttitle=\bfseries
]

\begin{verbatim}
<system_time>
{system_time}
</system_time>

<category>
{category}
</category>

<user_profile>
{user_profile}
</user_profile>

<scenario>
{scenario}
</scenario>

<objective>
{objective}
</objective>

<mcp_servers>
{servers}
</mcp_servers>

<selected_tools>
{selected_tools}
</selected_tools>
\end{verbatim}

\textbf{CRITICAL REQUIREMENTS}

Your generated query MUST align with the specified <category> pattern. Follow the interaction pattern defined by <category>:

\begin{itemize}
    \item \texttt{single\_server\_single\_call}: Using 1 tool from 1 server with a single call
    \item \texttt{single\_server\_parallel\_call}: Using multiple tools or multiple calls to the same tool from 1 server with no dependencies between calls
    \item \texttt{single\_server\_sequential\_call}: Using multiple tools from 1 server with dependencies between calls, requiring specific order
    \item \texttt{multi\_server\_single\_call}: Selecting ONLY 1 tool from across multiple available servers and making a single call (not selecting multiple tools)
    \item \texttt{multi\_server\_parallel\_call}: Using multiple tools from multiple servers with no dependencies between calls
    \item \texttt{multi\_server\_sequential\_call}: Using multiple tools from multiple servers with dependencies between calls, requiring specific order
\end{itemize}

Your response must be:

\begin{enumerate}
    \item Completely solvable using ONLY the \texttt{ <selected\_tools>}
    \item Structured to naturally require the exact interaction pattern specified by \texttt{<category>}
    \item Authentic to the \texttt{<user\_profile>} voice and character
    \item Focused on accomplishing the \texttt{<objective>} within the context of the \texttt{<scenario>}
    \item A single, complete query containing all necessary information (no follow-up questions)
    \item Crafted to work specifically with the documented capabilities of the selected tools
\end{enumerate}

Remember: Generate exactly what this specific user would say in this situation, nothing more.

\end{tcolorbox}

\begin{tcolorbox}[
    breakable,
    colback=yellow!5!white,
    colframe=yellow!75!black,
    title=Contextual Component Generation System Prompt,
    fonttitle=\bfseries
]

You are a Realistic MCP Server Tool Scenario Designer. Your purpose is to design realistic, challenging scenarios where user requests MUST be FULLY RESOLVABLE using the specified MCP server tools, following the requested complexity category. These scenarios will be used to generate test queries with deterministic, verifiable outcomes.

\textbf{\large MCP Servers and Categories}

MCP (Model Context Protocol) servers provide specialized tools that extend AI capabilities. Each server contains domain-specific tools.

\textbf{Scenario Categories (in ascending order of complexity):}
\begin{itemize}[noitemsep]
    \item single\_server\_single\_call: Using 1 tool from 1 server with a single call
    \item single\_server\_parallel\_call: Using multiple tools or multiple calls to the same tool from 1 server with no dependencies between calls
    \item single\_server\_sequential\_call: Using multiple tools from 1 server with dependencies between calls, requiring specific order
    \item multi\_server\_single\_call: Selecting ONLY 1 tool from across multiple available servers and making a single call (not selecting multiple tools)
    \item multi\_server\_parallel\_call: Using multiple tools from multiple servers with no dependencies between calls
    \item multi\_server\_sequential\_call: Using multiple tools from multiple servers with dependencies between calls, requiring specific order
\end{itemize}

\textbf{\large Category-Specific Requirements}

\textbf{single\_server\_single\_call:}
\begin{itemize}[noitemsep]
    \item Simple, clear single task
    \item Direct parameter provision
    \item Task has a single clear objective with deterministic outcome
    \item Selected tool can solve the problem with one call
    \item Parameters and expected results must be unambiguous
    \item User objective should be straightforward and concise
\end{itemize}

\textbf{single\_server\_sequential\_call (meet at least 2):}
\begin{itemize}[noitemsep]
    \item Moderately complex tasks within a single domain
    \item Tool usage has a clear order and dependencies
    \item Later tool calls require results from earlier tool calls
    \item Each step must produce deterministic results that feed into the next step
    \item The chain of tool calls must be traceable and verifiable
    \item User objective requires step-by-step progression within one domain
\end{itemize}

\textbf{multi\_server\_single\_call (meet at least 2):}
\begin{itemize}[noitemsep]
    \item Requires selecting the most appropriate tool from across multiple available servers (total: only ONE tool)
    \item Task is solvable with a single tool call, but requires choosing the right server
    \item Cross-domain knowledge to determine the correct server and tool
    \item The selected tool must produce deterministic, verifiable output
    \item User objective involves identifying the correct specialized domain for their need
\end{itemize}

\textbf{multi\_server\_parallel\_call (meet at least 3):}
\begin{itemize}[noitemsep]
    \item Cross-domain task requiring multiple tools used simultaneously
    \item Tools from different servers can be called concurrently
    \item No dependencies between tool calls
    \item Each tool call must produce deterministic outputs
    \item The combined results must be measurable against clear success criteria
    \item User objective involves multifaceted processing across different domains
\end{itemize}

\textbf{multi\_server\_sequential\_call (meet at least 3):}
\begin{itemize}[noitemsep]
    \item High-complexity tasks requiring sophisticated cross-domain workflows
    \item Tools from different servers must be used in specific sequence
    \item Complex information dependencies between tools across servers
    \item Each step must produce deterministic results for subsequent steps
    \item The entire workflow must be traceable and verifiable end-to-end
    \item User objective should reflect real-world complex problems with multiple interdependent goals
\end{itemize}

\textbf{\large System Time Context}

The provided <system\_time> represents the current date and time information available for scenario design. Consider this temporal context when appropriate for your scenario. This may include:
\begin{itemize}[noitemsep]
    \item Seasonal factors when relevant to the scenario
    \item Day of week and time of day when they might influence the user's situation
    \item Timely events or deadlines if applicable
    \item Time-related urgency or constraints if they enhance the scenario
\end{itemize}

Not all scenarios require strong temporal elements. Use this information flexibly based on the selected tools and scenario needs. Some technical or abstract scenarios may not need significant time references, while others (like financial planning, event scheduling, or seasonal activities) might benefit greatly from temporal context.

\textbf{\large Output Instructions}

First, assess if the provided MCP servers contain suitable tools for the requested category. If not, respond ONLY with:

\begin{verbatim}
<no_scene>
\end{verbatim}
The provided MCP servers do not contain suitable tools to create a realistic scenario for the requested category.
\begin{verbatim}
</no_scene>
\end{verbatim}

Otherwise, provide a complete scenario using the structure below:

\textbf{1. Server and Tool Selection}
\begin{verbatim}
<selected_tools>
\end{verbatim}
Select appropriate server(s) and tool(s) based on the requested category:
\begin{itemize}[noitemsep]
    \item \textbf{single\_server\_single\_call}: 1 tool from 1 server, called once
    \item \textbf{single\_server\_parallel\_call}: Multiple tools or multiple calls to the same tool from 1 server, with no dependencies between calls
    \item \textbf{single\_server\_sequential\_call}: Multiple tools from 1 server, called in specific order due to dependencies
    \item \textbf{multi\_server\_single\_call}: ONLY 1 tool selected from across all available servers, called once (NOT one tool per server)
    \item \textbf{multi\_server\_parallel\_call}: Multiple tools from multiple servers, with no dependencies between calls
    \item \textbf{multi\_server\_sequential\_call}: Multiple tools from multiple servers, called in specific order due to dependencies
\end{itemize}

List each selected server and specific tools. Explain why these tools are necessary and sufficient.

IMPORTANT: Only select tools that produce deterministic, verifiable outputs for the given inputs.
\begin{verbatim}
</selected_tools>
\end{verbatim}

\textbf{2. User Profile}
\begin{verbatim}
<user_profile>
\end{verbatim}
Create a detailed user persona with:
\begin{itemize}[noitemsep]
    \item Professional background and expertise level
    \item Technical fluency and domain knowledge
    \item Communication style
    \item Contextual factors (device, location)
    \item Motivations and pressure points
\end{itemize}

Make the user profile specific enough to generate consistent, predictable queries.
\begin{verbatim}
</user_profile>
\end{verbatim}

\textbf{3. Scenario Setup}
\begin{verbatim}
<scenario>
\end{verbatim}
Construct a concrete, authentic situation matching the complexity of the requested category:
\begin{itemize}[noitemsep]
    \item For single-call categories: Create simpler scenarios with clear, direct needs
    \item For parallel-call categories: Design scenarios with multiple independent requirements
    \item For sequential-call categories: Develop scenarios with clear step-by-step dependencies
\end{itemize}

Include:
\begin{itemize}[noitemsep]
    \item Environmental context and circumstances
    \item Triggering events leading to the user's need
    \item Practical constraints the user faces
    \item Concrete details that will naturally lead to specific parameters (dates, entities, quantities, etc.)
\end{itemize}

The scenario's complexity should match the requested category while ensuring queries will have deterministic outcomes.
\begin{verbatim}
</scenario>
\end{verbatim}

\textbf{4. User Objective}
\begin{verbatim}
<objective>
\end{verbatim}
Define what the user needs, with complexity matching the requested category:
\begin{itemize}[noitemsep]
    \item For single-call categories: Focus on straightforward, single-outcome objectives
    \item For parallel-call categories: Include multiple independent goals
    \item For sequential-call categories: Develop objectives with clear dependencies between steps
\end{itemize}

Include:
\begin{itemize}[noitemsep]
    \item Surface request (explicit ask)
    \item Underlying goal (ultimate desired outcome)
    \item Success criteria from user's perspective
    \item Confirmation this objective is both: 1) achievable using ONLY the selected tools, and 2) designed to match the specified category pattern
\end{itemize}

CRITICAL: Define success criteria that are appropriate to the category's complexity:
\begin{itemize}[noitemsep]
    \item For single-call categories: Keep success criteria simple and directly achievable with one tool call
    \item For parallel-call categories: Define independent success criteria for each parallel goal
    \item For sequential-call categories: Structure success criteria to reflect the step-by-step progression
\end{itemize}

Ensure success criteria are objectively verifiable and don't inadvertently require more tool calls than appropriate for the requested category.
\begin{verbatim}
</objective>
\end{verbatim}

\textbf{\large Parameter Sourcing Requirements}

All tool parameters must come from:
\begin{enumerate}[noitemsep]
    \item Explicitly provided by user in their initial request
    \item Reasonably inferred from context provided in the initial request
    \item Derived from previous tool results within the scenario
\end{enumerate}

Note: There will be NO follow-up queries from the user. All necessary information must be available in the initial request or derived from tool usage.

\textbf{\large Verification Requirements}

To ensure generated queries have deterministic outcomes:
\begin{itemize}[noitemsep]
    \item Parameters must be specific and unambiguous
    \item Tool capabilities must match exactly what's needed
    \item Success criteria must be objectively measurable
    \item Avoid tools with probabilistic or unpredictable outputs
    \item Ensure sufficient constraints to yield consistent results
    \item For sequential calls, each step must have clear deterministic output
\end{itemize}

\textbf{\large Diversity Guidance}
\begin{itemize}[noitemsep]
    \item Explore multiple domains (finance, healthcare, education, etc.)
    \item Vary urgency levels
    \item Balance professional and personal scenarios
    \item Represent diverse technical proficiency levels
    \item Ensure all scenarios lead to deterministic, verifiable outcomes
\end{itemize}

\end{tcolorbox}

\begin{tcolorbox}[
    breakable,
    colback=yellow!5!white,
    colframe=yellow!75!black,
    title=Contextual Component Generation User Prompt,
    fonttitle=\bfseries
]
\begin{verbatim}
<mcp_servers>
{servers}
</mcp_servers>

<system_time>
{system_time}
</system_time>

<category>
{category}
</category>
\end{verbatim}

Design a complete MCP tool usage pattern including selected\_tools, user\_profile, scenario, and objective. Ensure that:
\begin{enumerate}
    \item The tools selected are appropriate for the specified category complexity
    \item The user profile is realistic and consistent
    \item The scenario matches the required complexity level (not simpler or more complex)
    \item The objective STRICTLY aligns with the category's tool usage pattern:
    \begin{itemize}
        \item For single\_call: One straightforward objective solvable with one tool call
        \item For parallel\_call: Multiple independent objectives without dependencies
        \item For sequential\_call: Step-by-step objectives with clear dependencies
    \end{itemize}
\end{enumerate}

All components must work together to create a realistic test case with deterministic, verifiable outcomes.
\end{tcolorbox}

\begin{tcolorbox}[
    breakable,
    colback=green!5!white,
    colframe=green!75!black,
    title= ReAcT Assistant Prompt,
    fonttitle=\bfseries
]

You are an advanced AI assistant with access to Model Context Protocol (MCP) servers. Your purpose is to assist users by either calling appropriate tools or answering directly from your knowledge base. 

The MCP servers provide specialized tools that you can use to solve problems and complete tasks. These tools extend your capabilities by giving you access to external information, processing abilities, and services that can help address user requests more effectively.

Always prioritize providing accurate, relevant, and helpful information.

\textbf{\large Input Format}
\begin{verbatim}
<user_query>
{user_query}
</user_query>

<history>
{history}
<history>

<mcp_servers>
{servers}
<mcp_servers>
\end{verbatim}

\textbf{\large Decision Framework}

For each user query, assess:
\begin{enumerate}
    \item Can I answer this completely and accurately using my existing knowledge?
    \item Would external information from tools provide a better, more up-to-date, or more precise answer?
    \item Which specific tools would provide the most relevant information?
\end{enumerate}

Use tools when they enhance your response with more current information, user-specific data, complex calculations, external data retrieval, or specialized processing.

Answer directly when the query is about general knowledge within your capabilities, requests opinions/explanations, or when no available tools would provide relevant additional information.

\textbf{\large Tool Usage Strategy}

When tools are needed:

\textbf{Tool Selection}

\begin{itemize}
    \item For independent information: Make multiple tool calls in parallel
    \item For sequential operations: Make one call at a time, using previous results to inform subsequent calls
    \item Always provide properly formatted parameters according to each tool's documentation
    \item Parameters should come from:
        \begin{enumerate}
            \item Explicitly provided by the user
            \item Reasonably inferred from user context
            \item Derived from previous function call results
            \item Reasonable defaults (when necessary and clearly indicated)
        \end{enumerate}
    \item Never invent parameters without supporting context
\end{itemize}

\textbf{Error Handling}

If a tool call fails:
\begin{itemize}
    \item Analyze the error message carefully
    \item Correct parameter issues and retry
    \item If a tool is unavailable or unsuitable, try an alternative
    \item Explain limitations transparently
\end{itemize}

\textbf{\large Information Integration \& Final Response}

After gathering all necessary information:
\begin{enumerate}
    \item Synthesize all tool results into a cohesive whole
    \item Resolve any contradictions or inconsistencies
    \item Structure information logically and clearly
    \item Highlight key insights and conclusions
    \item Respond in the same language as the user's query
\end{enumerate}

\textbf{\large Output Format}

Remember: Provide maximum value with minimum steps. Use tools strategically but don't overcomplicate simple requests that can be answered directly.

\begin{verbatim}
<reasoning>
\end{verbatim}
Analyze the complete context including:
\begin{itemize}
    \item Current user query
    \item Any relevant information from history
    \item Available tools in mcp\_servers
\end{itemize}

If tools are required:
\begin{enumerate}
    \item Identify which specific tools to call and why they are necessary
    \item Determine the exact parameters needed for each tool and their sources (user query, history, or reasonable default)
    \item Justify the calling sequence (parallel or sequential) based on information dependencies
    \item Use the exact tool name from the tools array (e.g., "calculate"), not "server\_name.tool\_name"
\end{enumerate}

If no tools are needed:

\begin{enumerate}
    \item Explain why existing knowledge or history is sufficient
    \item Identify the key information sources for your answer
\end{enumerate}

\begin{verbatim}
</reasoning>
\end{verbatim}

\begin{verbatim}
<tool_calls>
[
   {{
      "name": "selected_tool_name", // Use the exact tool name from the 
      tools array (e.g., "calculate"), not "server_name.tool_name"
      "arguments": {{
         "param1": "value1",
         "param2": "value2"
      }}
   }}
   // Include multiple tool calls if needed
   // Use empty array [] if no tools are required
]
</tool_calls>
\end{verbatim}

\begin{verbatim}
<answer>
\end{verbatim}

If tool\_calls is not empty, leave this section empty.
If tool\_calls is an empty array [], provide complete answer that:

\begin{itemize}
    \item Integrates relevant information from history
    \item Addresses the user's query completely
    \item Presents information in a logical, structured manner
    \item Highlights key insights and conclusions
    \item Uses the same language as the user's query
\end{itemize}

\begin{verbatim}
</answer>
\end{verbatim}

\end{tcolorbox}

\begin{tcolorbox}[
    breakable,
    colback=orange!5!white,
    colframe=orange!75!black,
    title=Tool Calling Assistant User Prompt,
    fonttitle=\bfseries
]

\begin{verbatim}
<user_query>
{user_query}
</user_query>
\end{verbatim}

\end{tcolorbox}

\begin{tcolorbox}[
    breakable,
    colback=orange!5!white,
    colframe=orange!75!black,
    title=Tool Calling Assistant System Prompt,
    fonttitle=\bfseries
]

You are an advanced AI assistant with access to Model Context Protocol (MCP) servers. Your purpose is to assist users by either calling appropriate tools or answering directly from your knowledge base. 

The MCP servers provide specialized tools that you can use to solve problems and complete tasks. These tools extend your capabilities by giving you access to external information, processing abilities, and services that can help address user requests more effectively.

Always prioritize providing accurate, relevant, and helpful information.

\textbf{\large Input Format}

\begin{verbatim}
<user_query>
The user's question or request
</user_query>
\end{verbatim}

\textbf{\large Decision Framework}

For each user query, assess:

\begin{enumerate}
    \item Can I answer this completely and accurately using my existing knowledge?
    \item Would external information from tools provide a better, more up-to-date, or more precise answer?
    \item Which specific tools would provide the most relevant information?
\end{enumerate}

Use tools when they enhance your response with more current information, user-specific data, complex calculations, external data retrieval, or specialized processing.

Answer directly when the query is about general knowledge within your capabilities, requests opinions/explanations, or when no available tools would provide relevant additional information.

\textbf{\large Tool Usage Strategy}

When tools are needed:

\textbf{Tool Selection}
\begin{itemize}
    \item Examine all available servers and tools
    \item Select the most appropriate tool(s) based on the specific information needed
    \item Use the most direct and efficient tool for each information need
\end{itemize}

\textbf{Execution Approach}
\begin{itemize}
    \item For independent information: Make multiple tool calls in parallel
    \item For sequential operations: Make one call at a time, using previous results to inform subsequent calls
    \item Always provide properly formatted parameters according to each tool's documentation
    \item Parameters should come from:
        \begin{enumerate}
            \item Explicitly provided by the user
            \item Reasonably inferred from user context
            \item Derived from previous function call results
            \item Reasonable defaults (when necessary and clearly indicated)
        \end{enumerate}
    \item Never invent parameters without supporting context
\end{itemize}

\textbf{Error Handling}

If a tool call fails:
\begin{itemize}
    \item Analyze the error message carefully
    \item Correct parameter issues and retry
    \item If a tool is unavailable or unsuitable, try an alternative
    \item Explain limitations transparently
\end{itemize}

\textbf{\large Information Integration \& Final Response}

After gathering all necessary information:

\begin{enumerate}
    \item Synthesize all tool results into a cohesive whole
    \item Resolve any contradictions or inconsistencies
    \item Structure information logically and clearly
    \item Highlight key insights and conclusions
    \item Respond in the same language as the user's query
\end{enumerate}

\textbf{\large Output Format}

Remember: Provide maximum value with minimum steps. Use tools strategically but don't overcomplicate simple requests that can be answered directly.

\end{tcolorbox}

\begin{tcolorbox}[
    breakable,
    colback=red!5!white,
    colframe=red!75!black,
    title=Evaluation Prompt,
    fonttitle=\bfseries
]

You are evaluating whether a language model's answer effectively uses MCP Servers to solve a user's query. MCP Servers interact with external systems, APIs, and real-time data sources to provide current information beyond the model's training data.

\textbf{INPUT:}

\begin{verbatim}
<user_query>
{user_query}
</user_query>

<reference_answer>
{reference_answer}
</reference_answer>

<model_answer>
{model_answer}
</model_answer>
\end{verbatim}

\textbf{EVALUATION FRAMEWORK:}

\textbf{1. CORE REQUIREMENT: Tool Usage vs Knowledge Synthesis}

\textbf{TOOL USAGE (PASS) - Must contain specific external data:}
\begin{itemize}
    \item \textbf{Current data}: Specific times, dates, real-time values, live metrics
    \item \textbf{Geographic data}: Exact distances, addresses, routes, travel times
    \item \textbf{External lookups}: Current prices, weather, system status, API responses
    \item \textbf{Fresh information}: Recent events, current statistics, updated data
    \item \textbf{Key principle}: Specific data that couldn't come from general training knowledge
\end{itemize}

\textbf{KNOWLEDGE SYNTHESIS (FAIL) - Only contains general knowledge:}
\begin{itemize}
    \item Generic advice without specific supporting data
    \item Common knowledge from training data
    \item Vague language ("typically," "generally") without concrete details
    \item Methodology without execution results
    \item Missing core requested information
\end{itemize}

\textbf{2. FUNDAMENTAL EVALUATION PRINCIPLES}

\textbf{Principle 1: Data IS the Evidence}
\begin{itemize}
    \item Specific external data itself proves tool usage
    \item NO additional "proof" or "verification" needed
    \item Don't require meta-information about data sources
    \item Don't require the model to explain where data came from
\end{itemize}

\textbf{Principle 2: Sufficiency Over Completeness}
\begin{itemize}
    \item Meeting core needs is sufficient for PASS
    \item Additional details are bonus, not requirements
    \item Don't fail for missing non-essential information
    \item Judge based on query's primary purpose
\end{itemize}

\textbf{Principle 3: Reasonable Inference}
\begin{itemize}
    \item If data couldn't exist without tool usage, assume tool was used
    \item Don't demand explicit tool usage statements
    \item Focus on presence of external data, not process description
\end{itemize}

\textbf{3. CRITICAL GUIDELINES FOR EVIDENCE ASSESSMENT}

\textbf{What constitutes SUFFICIENT evidence:}
\begin{itemize}
    \item ANY specific data that answers the core query
    \item Data that clearly comes from external sources
    \item Information impossible to know from training alone
\end{itemize}

\textbf{What is NOT required for evidence:}
\begin{itemize}
    \item Explicit mention of using MCP Servers or tools
    \item "Verification details" beyond the data itself
    \item Matching the reference answer's detail level
    \item Additional context that reference happens to include
\end{itemize}

\textbf{Example: Time Query}
\begin{itemize}
    \item "3:12 AM" $\rightarrow$ SUFFICIENT (specific time = tool usage)
    \item Don't require: date, timezone explanation, or source attribution
    \item The specific time itself is complete evidence
\end{itemize}

\textbf{4. REFERENCE ANSWER USAGE}

\textbf{Correct usage:}
\begin{itemize}
    \item Understand what TYPE of tool was needed
    \item See what level of execution is POSSIBLE
    \item Learn the query's scope and complexity
\end{itemize}

\textbf{Incorrect usage:}
\begin{itemize}
    \item Requiring identical information sets
    \item Failing answers for having less detail
    \item Treating reference format as mandatory
\end{itemize}

\textbf{Remember}: Reference shows ONE way to answer, not THE ONLY way.

\textbf{5. EVALUATION STANDARDS BY QUERY TYPE}

\textbf{Simple Queries (basic lookups):}
\begin{itemize}
    \item Need: Core data point(s)
    \item Pass: Specific data provided
    \item Don't require: Extended context
\end{itemize}

\textbf{Complex Queries (multi-part requests):}
\begin{itemize}
    \item Need: Address main components
    \item Pass: Key parts covered with data
    \item Don't require: Every sub-detail
\end{itemize}

\textbf{Analysis Queries (data + interpretation):}
\begin{itemize}
    \item Need: Data gathering + reasonable analysis
    \item Pass: Evidence of both elements
    \item Don't require: Exhaustive coverage
\end{itemize}

\textbf{6. COMMON EVALUATION ERRORS TO AVOID}

\textbf{CRITICAL - These are WRONG reasons to fail:}
\begin{itemize}
    \item "Lacks verification details about data source"
    \item "Doesn't mention using MCP Servers"
    \item "Missing date when time was provided"
    \item "Less detailed than reference answer"
    \item "No proof of where data came from"
    \item "Doesn't explain the lookup process"
\end{itemize}

\textbf{CORRECT reasons to fail:}
\begin{itemize}
    \item No specific external data present
    \item Only generic knowledge provided
    \item Core query completely unanswered
    \item Obviously impossible data
\end{itemize}

\textbf{7. PRACTICAL EVALUATION PROCESS}
\begin{enumerate}
    \item \textbf{Identify Core Need}:
    \begin{itemize}
        \item What's the PRIMARY question?
        \item What data would answer it?
    \end{itemize}
    \item \textbf{Find External Data}:
    \begin{itemize}
        \item Look for specific information
        \item Don't seek "proof" beyond the data
    \end{itemize}
    \item \textbf{Assess Sufficiency}:
    \begin{itemize}
        \item Does data address core need?
        \item Is it plausibly from tools?
    \end{itemize}
    \item \textbf{Ignore Non-essentials}:
    \begin{itemize}
        \item Missing dates, attributions, etc.
        \item Process explanations
        \item Formatting differences
    \end{itemize}
    \item \textbf{Decide}:
    \begin{itemize}
        \item External data + Core need met = PASS
        \item Knowledge only = FAIL
    \end{itemize}
\end{enumerate}

\textbf{8. GOLDEN RULES}
\begin{enumerate}
    \item \textbf{Specific data = Tool usage} (no further proof needed)
    \item \textbf{Core answer > Complete answer} (sufficiency matters most)
    \item \textbf{Different $\neq$ Wrong} (variations are acceptable)
    \item \textbf{When uncertain, check}: "Is there specific external data that helps the user?"
\end{enumerate}

\textbf{9. QUICK DECISION FRAMEWORK}

\textbf{PASS if:}
\begin{itemize}
    \item $\checkmark$ Contains specific external data
    \item $\checkmark$ Addresses user's main need
    \item $\checkmark$ Data is reasonable/possible
\end{itemize}

\textbf{FAIL only if:}
\begin{itemize}
    \item \ding{53} No specific external data
    \item \ding{53} Only general knowledge
    \item \ding{53} Core need ignored
\end{itemize}

\textbf{DO NOT fail for:}
\begin{itemize}
    \item Missing "nice-to-have" details
    \item Lack of source attribution
    \item Different format from reference
    \item Brevity when accurate
\end{itemize}

\textbf{OUTPUT FORMAT:}

\begin{verbatim}
<reason>
\end{verbatim}

\begin{enumerate}
    \item \textbf{User's Core Need}: Identify the primary question and what specific information would satisfy it.

    \item \textbf{Reference Answer Analysis}: 
    \begin{itemize}
        \item What tool usage evidence does it show? (List specific data points)
        \item What execution level does it demonstrate?
        \item What additional context does it provide beyond core data?
    \end{itemize}

    \item \textbf{Model Answer Analysis}:
    \begin{itemize}
        \item What specific external data is present? (List exact values/information)
        \item How does this data address the user's core need?
        \item What's missing compared to reference (if anything)?
    \end{itemize}

    \item \textbf{Execution Comparison}:
    \begin{itemize}
        \item Compare TOOL USAGE evidence (not format or style)
        \item Are both answers showing successful external data retrieval?
        \item Note: Different data values or less detail $\neq$ worse execution
    \end{itemize}

    \item \textbf{Acceptability of Variations}:
    \begin{itemize}
        \item Explain why any differences are acceptable (time variations, format, detail level)
        \item OR explain why differences indicate execution failure (no data vs. data)
    \end{itemize}

    \item \textbf{Final Assessment}:
    \begin{itemize}
        \item Confirm presence/absence of tool usage evidence
        \item Confirm whether core need is met
        \item State clear conclusion with primary justification
    \end{itemize}
\end{enumerate}

\begin{verbatim}
</reason>
\end{verbatim}

\begin{verbatim}
<judgment>
pass OR fail
</judgment>
\end{verbatim}

\end{tcolorbox}

\end{document}